\documentclass[english]{article}
\usepackage[T1]{fontenc}
\usepackage[utf8]{inputenc}
\usepackage{geometry}
\usepackage{color}
\usepackage{babel}
\usepackage{float}
\usepackage{textcomp}
\usepackage{amsmath}
\usepackage{amsthm}
\usepackage{setspace}
\usepackage[unicode=true,pdfusetitle,
 bookmarks=true,bookmarksnumbered=false,bookmarksopen=false,
 breaklinks=false,pdfborder={0 0 1},backref=false,colorlinks=false]
 {hyperref}

\makeatletter
\theoremstyle{remark}
\newtheorem{summary}{\protect\summaryname}
\theoremstyle{definition}
\newtheorem{defn}{\protect\definitionname}
\theoremstyle{plain}
\newtheorem{assumption}{\protect\assumptionname}
\theoremstyle{plain}
\newtheorem{criterion}{\protect\criterionname}
\theoremstyle{definition}
\newtheorem{sol}{\protect\solutionname}
\theoremstyle{plain}
\newtheorem{lem}{\protect\lemmaname}

\makeatother

\providecommand{\assumptionname}{Assumption}
\providecommand{\criterionname}{Criterion}
\providecommand{\definitionname}{Definition}
\providecommand{\lemmaname}{Lemma}
\providecommand{\solutionname}{Solution}
\providecommand{\summaryname}{Summary}

\begin{document}
\begin{doublespace}
\begin{center}
\textbf{\textcolor{black}{\Large{}Artificial Intelligence: A Child's
Play}}{\Large\par}
\par\end{center}

\begin{center}
\textbf{Ravi Kashyap (ravi.kashyap@stern.nyu.edu)}
\par\end{center}

\begin{center}
\textbf{City University of Hong Kong}\footnote{\begin{doublespace}
Numerous seminar participants suggested ways to improve the manuscript.
The views and opinions expressed in this article, along with any mistakes,
are mine alone and do not necessarily reflect the official policy,
or, position of either of my affiliations, or, any other agency. Dr.
Yong Wang, Dr. Isabel Yan, Dr. Vikas Kakkar, Dr. Fred Kwan, Dr. William
Case, Dr. Srikant Marakani, Dr. Qiang Zhang, Dr. Costel Andonie, Dr.
Jeff Hong, Dr. Guangwu Liu, Dr. Humphrey Tung and Dr. Xu Han at the
City University of Hong Kong; Dr. Richard Sylla, Dr. Adam Brandenburger,
Dr. Richard Freedman, Dr. Robert Engle, Prof. Larry Zicklin, Prof.
Seth Freeman, Dr. Laura Veldkamp, Dr. Ignacio Esponda at New York
University; Dr. Liam Lenten at La Trobe University; and Dr. Paul Joseph,
Dr. M. N. Neelakantan, Dr. V. K. Govindan, Dr. Moiuddin Kutty, Dr.
M.P. Sebastian, Mr. Murali Krishnan at the National Institute of Technology
Calicut provided valuable suggestions to explore and where possible
apply cross disciplinary techniques.
\end{doublespace}
}\textbf{ / SP Jain School of Global Management / Singapore University
of Social Sciences}
\par\end{center}

\begin{center}
\begin{center}
\today
\par\end{center}
\par\end{center}
\end{doublespace}

\begin{onehalfspace}
\begin{center}
\par\end{center}

\begin{center}
Physics Subject Headings: Complex Systems; Neuroscience; Statistical
Physics; Stochastic Processes; Machine Learning; Computational Techniques;
Ecology \& Evolution; Scientific reasoning \& problem solving
\par\end{center}

\begin{center}
\par\end{center}

\begin{center}
American Psychological Association Classification Codes: 4120 Artificial
Intelligence \& Expert Systems; 4100 Intelligent Systems; 2343 Learning
\& Memory; 2340 Cognitive Processes; 2630 Philosophy
\par\end{center}

\begin{center}
Journal of Economic Literature Codes: D83 Learning, Belief; C45 Neural
Networks \& Related Topics; D81 Criteria for Decision-Making under
Risk \& Uncertainty; D87 Neuro-Economics; C01 Econometrics
\par\end{center}

\begin{center}
Association for Computing Machinery Classification System: I.2.0:
General Artificial Intelligence; I.2.6: Learning; I.2.8: Problem Solving;
F.4.3: Formal Languages; G.3: Probability \& Statistics
\par\end{center}
\end{onehalfspace}

\begin{doublespace}
\begin{center}
\textbf{\textcolor{blue}{\href{https://doi.org/10.1016/j.techfore.2020.120555}{Edited Version: Kashyap, R. (2021).  Artificial Intelligence: A Child's Play.  Technological Forecasting \& Social Change,  166(5),  120555. }}}
\par\end{center}
\end{doublespace}

\begin{doublespace}
\tableofcontents{}
\end{doublespace}
\begin{doublespace}

\section{Abstract}
\end{doublespace}

\begin{doublespace}
We discuss the objectives of any endeavor in creating artificial intelligence,
AI, and provide a possible alternative. Intelligence might be an unintended
consequence of curiosity left to roam free, best exemplified by a
frolicking infant. This suggests that our attempts at AI could have
been misguided. What we actually need to strive for can be termed
artificial curiosity, AC, and intelligence happens as a consequence
of those efforts. For this unintentional yet welcome aftereffect to
set in a foundational list of guiding principles needs to be present.
We start with the intuition for this line of reasoning and formalize
it with a series of definitions, assumptions, ingredients, models
and iterative improvements that will be necessary to make the incubation
of intelligence a reality. Our discussion provides conceptual modifications
to the Turing Test and to Searle's Chinese room argument. We discuss
the future implications for society as AI becomes an integral part
of life.

We provide a road-map for creating intelligence with the technical
parts relegated to the appendix so that the article is accessible
to a wide audience. The central techniques in our formal approach
to creating intelligence draw upon tools and concepts widely used
in physics, cognitive science, psychology, evolutionary biology, statistics,
linguistics, communication systems, pattern recognition, marketing,
economics, finance, information science and computational theory highlighting
that solutions for creating artificial intelligence have to transcend
the artificial barriers between various fields and be highly multi-disciplinary.\pagebreak{}
\end{doublespace}
\begin{doublespace}

\section{\label{sec:The-Benchmark-for}The Benchmark for Brainpower}
\end{doublespace}

\begin{doublespace}
As a first step, we recognize that one possible categorization of
different fields can be done by the set of questions a particular
field attempts to answer. We are the creators of different disciplines
but not the creators of the world (based on our present state of understanding)
in which these fields need to operate. Hence, the answers to the questions
posed by any domain can come from anywhere or from phenomena studied
under a combination of many other disciplines. This implies that the
answers to the questions posed under the realm of artificial intelligence
(AI) can come from seemingly diverse subjects such as: physics, biology,
psychology, mathematics, chemistry, marketing, engineering, economics,
literature, theater, music and so on. 

This suggests that we might be better off identifying ourselves with
problems and solutions, which tacitly confers upon us the title Problem
Solvers, instead of calling ourselves physicists, biologists, psychologists,
mathematicians, engineers, chemists, marketing experts, economists,
and the like. It would not be entirely incorrect to state that the
majority of our attempts at solving problems start with posing well
defined questions and finding corresponding answers. As we linger
on the topic of Questions \& Answers, Q\&A, we need to be cognizant
that any answer we wish to seek would depend on some Definitions and
Assumptions, D\&A. But it is absolutely essential to keep in mind
that if we change those D\&A we might get different Q\&A (End-note
\ref{enu:Changing-D=000026A-which}). 

Hence in later sections, we start with a definition of intelligence.
We then highlight the assumptions and criteria, under which we attempt
to seek the answers, for questions related to the creation of intelligence.
The structure of the paper based on this flow of arguments consists
of Definitions and Assumptions, which act as the foundation, upon
which we provide Solutions and Criteria that supply testable ideas
and instruments for practical applications.

For simplicity, and to be specific, we could confine the creation
of intelligence outside the confines of biological organisms. But
it will become clear later on that many parts of our discussion apply
to the goals of increasing intelligence within biological organisms
as well. Not to mention, as discussed above, a suitable definition
could render biological and non-biological organisms under the same
category of sentient beings.

The problem of designing intelligence artificially can be a rather
trivial task depending on which organism's brainpower acts as our
gold standard. A simple criterion for the problem of creating artificial
intelligence would make it a child's play, or a very straightforward
task. What we also mean by a child's play is that children are still
playing even as they are learning. Perhaps, the real challenge is
to replicate the curiosity and the level of engagement an infant displays.
We discuss the objectives of any endeavor in creating artificial intelligence
(Sections \ref{subsec:Questionably-Simple-Yet}; \ref{subsec:Intelligence-for-What};
\ref{subsec:Minds-versus-Machines}), AI, and provide a possible alternative.
\end{doublespace}
\begin{summary}
\begin{doublespace}
\textit{Intelligence might be an unintended consequence of curiosity
left to roam free, best exemplified by a frolicking infant (Section
\ref{subsec:Becoming-Smarter-than}).} \textit{This suggests that
our attempts at AI could have been misguided. What we actually need
to strive for can be termed artificial curiosity, AC, and intelligence
happens as a consequence of those efforts. }
\end{doublespace}
\end{summary}
\begin{doublespace}
For this unintentional yet welcome aftereffect to set in, a foundational
list of guiding principles needs to be present (Section \ref{sec:A-Journey-to}).
We consider what these essential doctrines might be. We discuss why
their establishment is required to form connections, possibly growing,
between a knowledge store that has been built up and new pieces of
information that curiosity will bring back. As more findings are acquired
and more bonds are fermented we need a way to periodically reduce
the amount of data. In a sense, it is important to capture the critical
characteristics of what has been accumulated or to produce a summary
of what has been gathered. Curiosity helps to collect material that
can be useful for decision making, but those constituents have to
marshaled successfully towards a decision making goal.

We start with the intuition for this line of reasoning and formalize
it with a series of definitions, assumptions, ingredients, models
and iterative improvements that will be necessary to make the incubation
of intelligence a reality. Section (\ref{sec:A-Road-map-for}) provides
a road-map for creating intelligence. The technical parts have been
relegated to Appendix (\ref{sec:From-Words-to-Symbols}), which has
the mathematical elements for creating intelligence, and can be incorporated
into suitable algorithms or machine learning systems. This approach
ensures that the paper is written in a non-technical language, to
facilitate understanding by a wide audience, making it accessible
to almost anyone interested in AI. While the appendices provide sufficient
rigor to enable technological implementations of the ideas. Our discussion
provides conceptual modifications to the Turing Test and to Searle's
Chinese room argument (Sections \ref{subsec:Acing-the-Turing}; \ref{subsec:Imitation-in-the};
\ref{subsec:Mexican-Chihuahua-solving}). 

The central techniques in our formal approach to creating intelligence
draw upon tools and concepts widely used in physics, cognitive science,
psychology, evolutionary biology, statistics, linguistics, communication
systems, pattern recognition, marketing, economics, finance, information
science and computational theory. This highlights that solutions \textit{``for
creating artificial intelligence''} have to \textit{``transcend
the artificial barriers''} between various fields and be highly multi-disciplinary.
In addition, since every field will benefit from increased intelligence,
the question of creating intelligence belongs to every discipline.
We consider many unintended consequences, one of the main themes of
this paper, in the quest for intelligence and the future implications
for society as AI becomes an integral part of life.
\end{doublespace}
\begin{doublespace}

\subsection{\label{subsec:Questionably-Simple-Yet}Questionably Simple Yet Complex
Benchmark}
\end{doublespace}

\begin{doublespace}
As we embark on the journey to apply the knowledge from other fields
to AI we need to be aware that artificial intelligence is ``Simply
Too Complex''. This is because through time AI has just been about
beating a benchmark. The complications are mainly to select the right
standards to compete with. This problem is compounded due to the fact
that nobody really knows what is intelligence, especially when considering
artificial systems. (Legg \& Hutter 2007) take a number of well known
informal definitions of human intelligence and extract their essential
features, which are then mathematically formalized to produce a general
measure of intelligence. 

Intelligence is defined and approached in many ways. To facilitate
a reference point for the rest of the article we define intelligence
as below. This additional attempt perhaps compounds the prevailing
confusion. Though Section (\ref{subsec:I-Don't-Know,}) clarifies
why this might still be a positive outcome.
\end{doublespace}
\begin{defn}
\begin{doublespace}
\textit{\label{def:Intelligence-is-the}Intelligence is the ability
to connect elements of previously attained information to effect a
decision. Nothing lasts forever and hence no decision is good forever.
But the longer a decision serves its purpose, the greater the intelligence
of the agent making that decision.}
\end{doublespace}
\end{defn}
\begin{doublespace}
With this definition it is implied that even a very intelligent decision
(such as in a game of chess), that falls short of meeting its objective
when it has to counter a decision with greater intelligence, might
fail the benchmark. Intelligence is with respect to the situation
and its demands highlighting that we can only make relative comparisons
of intelligence. Many artificial systems have obtained increasing
levels of sophistication, but the real test is when they can continue
to counter, or outperform, situations with greater requirements. Another
example is with regards to autonomous driving where remarkable progress
has been made over the years. If autonomous vehicles continue to have
greater number of accidents (measured using percentages or other statistical
metrics) as compared to vehicles with human drivers, the intelligence
that has been created is not sufficient. Intelligence is not about
simply being intelligent but it is about being intelligent enough.

It should also become clear that intelligence requires the ability
to collect pieces of information and to connect them towards a decision
making goal. Since decisions are not going to be valid indefinitely
we need to continue to use these abilities. This could be stated explicitly
as an assumption. The notion of connecting known pieces of information
to obtain a new combination is extensively studied under the heading
of innovation and is acknowledged as the key process behind creativity
(Young 1965). We further restrict our discussion to the sub class
of living organisms termed the homo sapiens and the agents being created
by them, using computers and related software, to possess intelligence.
We specifically check how this definition can be applied in the context
of the Turing Test in Section (\ref{subsec:Acing-the-Turing}). 

With no disrespect to any adults, it would not be entirely wrong to
label children as better and faster learners than adults. (Holt 2017)
shows that in most situations our minds work best when we use them
in a certain way. He suggests that young children tend to learn better
than grownups (and better than they themselves will when they are
older) because they use their minds in a special way, which is a style
of learning that fits their present condition.

(Russell \& Norvig 1995) is a comprehensive discussion of the concept
of an intelligent agent. (Wooldridge \& Jennings 1995) discuss the
most important theoretical and practical issues associated with the
design and construction of intelligent agents. They divide these issues
into three areas (clearly most divisions cannot completely rule out
overlap between the components).
\end{doublespace}
\begin{enumerate}
\begin{doublespace}
\item Agent theory is concerned with the question of what an agent is, and
the use of mathematical formalisms for representing and reasoning
about the properties of agents.
\item Agent architectures is about the software engineering models of agents.
This area is primarily concerned with the problem of designing software
or hardware systems that will satisfy the properties specified by
agent theorists.
\item Agent languages are software systems for programming and experimenting
with agents. These languages may embody principles proposed by theorists.
\end{doublespace}
\end{enumerate}
\begin{doublespace}
Neural networks are one approach to artificial intelligence (AI) that
are modeled on the brain (Haykin 2004; Haykin 2009; Castelvecchi 2016).
These systems, loosely inspired by the densely interconnected neurons
of the brain, mimic human learning by changing the strength of simulated
neural connections on the basis of experience. Unfortunately, such
networks are also as opaque as the brain though they promised to be
better than standard algorithms at dealing with complex real-world
situations. Instead of storing what they have learned in a neat block
of digital memory, they diffuse the information in a way that is exceedingly
difficult to decipher.

Deep learning is a three-decade-old technique in which massive amounts
of data and processing power help computers to crack messy problems
that humans solve almost intuitively, from recognizing faces to understanding
language (Jones 2014; LeCun, Bengio \& Hinton 2015). Such methods
fall under a broader category termed machine learning, which aims
to program computers to use example data or past experience to solve
a given problem (Alpaydin 2014). Using the data to decipher patterns
is also known as training the system. Deep learning models are built
as artificial neural networks and use a cascade of multiple layers
of nonlinear processing units, allowing computational models to learn
representations of data with multiple levels of abstraction. Each
successive layer uses the output from the previous layer as input.
(Deng \& Yu 2014) is overview of deep learning methodologies and their
applications to a variety of information processing tasks. (Schmidhuber
2015) is a comprehensive survey about deep learning in neural networks.
\end{doublespace}
\begin{doublespace}

\subsection{\label{subsec:Intelligence-for-What}Intelligence for What Sake?}
\end{doublespace}

\begin{doublespace}
To be precise, this section is not about creating intelligence to
barter for the Japanese drink, sake. Though that seems like a wise
exchange and might have been done many times before. 

A central aspect of our lives is uncertainty and our struggle to overcome
it. Over the years, it seems that we have found ways to understand
the uncertainty in the natural world by postulating numerous physical
laws. The majority of the predictions in the physical world hold under
a fairly robust set of circumstances and cannot be influenced by the
person making the observation. These predictions stay unaffected if
more people become aware of such a possibility. In the social sciences,
the situation is exactly the contrary. (Popper 2002) gave a critique
and warned of the dangers of historical prediction in social systems.

We need intelligent decision making because of the uncertainty in
the world we live in. Hence perhaps, the one central theme in this
entire article is Uncertainty. The dynamic nature of the social sciences,
where changes can be observed and decisions can be taken by participants
to influence the system, means that along with better models and predictive
technologies, predictions need to be continuously revised. And yet
unintended consequences set in and as long as participants are free
to observe the results and modify their actions, this effect will
persist (Kashyap 2016). 

A hall mark of the social sciences is the lack of objectivity. Here
we assert that objectivity is with respect to comparisons done by
different participants and that a comparison is a precursor to a decision.
\end{doublespace}
\begin{assumption}
\begin{doublespace}
Despite the several advances in the social sciences, we have yet to
discover an objective measuring stick for comparison, a so called,
True Comparison Theory, which can be an aid for arriving at objective
decisions. 
\end{doublespace}
\end{assumption}
\begin{doublespace}
For our present purposes the lack of such an objective measure means
that the difference in comparisons, as assessed by different participants,
can effect different decisions under the same set of circumstances.
Hence, despite all the uncertainty in the social sciences, the one
thing we can be almost certain about is the subjectivity in all decision
making. This lack of an objective measure for comparisons makes people
react at varying degrees and at varying speeds as they make their
subjective decisions. A decision gives rise to an action and subjectivity
in the comparison means differing decisions and hence unpredictable
actions. This inability to make consistent predictions in the social
sciences explains the growing trend towards collecting more information
across the entire cycle of comparisons, decisions and actions. The
goal being better comprehension and deciphering of the decision process
and the subsequent actions.

Another feature of the social sciences is that the actions of participants
affects the state of the system. This effects a state transfer which
perpetuates another merry-go-round of comparisons, decisions and actions
from the participants involved. This means, more the participants,
more the changes to the system, more the actions and more the information
that is generated to be gathered. Hence perhaps, an unintended consequence
of the recent developments in technology has been to increase the
complexity in our lives in many ways.

(Simon 1962) points out that any attempt to seek properties common
to many sorts of complex systems (physical, biological or social),
would lead to a theory of hierarchy since a large proportion of complex
systems observed in nature exhibit hierarchic structure. That is a
complex system is composed of subsystems that in turn have their own
subsystems, and so on. 

This might hold a clue to the marvel that our minds perform, abstracting
away from the dots that make up a picture, to fully visualizing the
image that seems far removed from the pieces that give form and meaning
to it. To help us gain a better understanding of the relationships
between different elements of information we use a metric built from
smaller parts (Section \ref{sec:From-Words-to-Symbols}) that gives
optimal benefits when seen from a higher level. Contrary to what conventional
big picture conversations suggest, as the spectator steps back and
the distance from the picture increases, the image becomes smaller
yet clearer.

(McManus \& Hastings 2005) clarify the wide range of uncertainties
that affect complex engineering systems. They present a framework
to understand the risks and opportunities uncertainties create and
the strategies system designers can use to mitigate or take advantage
of them. (Keynes 1937; 1971; 1973) contends that it is generally impossible,
even in probabilistic terms, to evaluate the future outcomes of all
possible current actions. (Lawson 1985) argues that the Keynesian
view on uncertainty, far from being innocuous or destructive of economic
analysis in general, could be potentially fruitful by giving rise
to research programs incorporating, amongst other things, a view of
rational behavior under uncertainty. 

These viewpoints hold many lessons for AI designers and could be instructive
for researchers looking to create methods to compare and build complex
systems, keeping in mind the caveats of dynamic social systems.
\end{doublespace}
\begin{doublespace}

\subsection{\label{subsec:Minds-versus-Machines}Minds versus Machines}
\end{doublespace}

\begin{doublespace}
We currently lack a proper understanding of how, in some instances,
our brains or minds (right now, it seems, we don't know the difference!)
make the leap of learning from information to knowledge to wisdom.
(Mill 1829; Mazur 2015) have an excellent account of learning and
behavior. Intellect might be a byproduct of Inquisitiveness, demonstrating
another instance of an unintended yet welcome consequence (Kashyap
2016). If ignorance is bliss, intrusion might just be the opposite
and bring misery. As the saying goes, Curiosity Terminated the Cat
and the movie Terminator should tell us about other unintended consequences
that might pop up in the AI adventure (Cameron \& Wisher 1991).

This brings up the question of Art and Science in the creation of
AI (and everything else in life?), which are more related than we
probably realize. Art is Science that we don't know about. Science
is Art restricted to a set of symbols governed by a growing number
of rules (End-note \ref{enu:Art-Science}). While the similarities
between art and science should give us hope, we need to face the realities
of the situation. Right now, arguably in most cases, we (including
computers and intelligent machines?) can barely make the jump from
the information to the knowledge stage even with the use of cutting
(bleeding?) edge technology and tools. This exemplifies three things: 
\end{doublespace}
\begin{enumerate}
\begin{doublespace}
\item We are still in the information age. As a route to establishing this,
consider the following argument. Information is Hidden. Knowledge
is Exchanged or Bartered. Wisdom is Dispersed. Surely, we are still
in the Information Age since a disproportionate amount of our actions
are geared towards accumulating unique data-sets for the sole benefits
of the accumulators. It is reassuring that this trend might get reversed
since datasets, software and other artifacts are being shared more
than before. The creation of many online platforms has been a blessing
to the members who benefit from discussion forums, direct messages
and other forms of collaboration (Phillips, Lin, Schifter \& Folse
2019). This might help us to accelerate to the next stages.
\item Automating the movement to a higher level of learning, which is necessary
for dealing with certain doses of uncertainty, is still far away. 
\item Some of us missed the memo that the best of humanity are actually
robots in disguise, living amongst us.
\end{doublespace}
\end{enumerate}
\begin{doublespace}
Hence, it is not Mind versus Machine. Not even, Man versus Machine
or MAN vs MAC, in short. Not even MAN and MAC against the MPC, Microsoft
Personal Computer (Freiberger \& Swaine 1999; Garland 1977; Campbell-Kelly
2001; Manes \& Andrews 1993; Carlton \& Annotations-Kawasaki 1997;
Wonglimpiyarat 2012; Corcoran, Coughlin \& Wozniak 2016; End-note
\ref{enu:MAC vs MPC})? It is MAN, MAC and the MPC against increasing
complexity. For the underlying concepts on which modern computers
are built and what the future holds see: (Davis 2011; Perrier, Sipper
\& Zahnd 1996; Denning 2005; Amir, etal 2014; Thompson, etal 2016;
End-notes \ref{enu:Universal Computing Machine}; \ref{enu:Computer}).
Also in scope are other computing platforms from the past, present
and the future: (Williams 1997; Ifrah etal 2000; Leuenberger \& Loss
2001; Ceruzzi 2003; Armbrust, etal 2010; Zhang, Cheng \& Boutaba 2010;
End-note \ref{enu:History Computing}). 

This increasing complexity and information explosion is perhaps due
to the increasing number of complex actions perpetrated by the actors
that comprise the social system. The human mind will be obsolete if
machines can fully manage society and we might have bigger problems
on our hands than who is taking care of things. We need, and will
continue to need, massive computing power and all the intelligence
we can create to mostly separate the signal from the noise. In this
age of (Too Much) Information, it is imperative for Man and Machine
to work together to uncover nuggets of knowledge from buckets of nonsense.
\end{doublespace}
\begin{doublespace}

\subsection{\label{subsec:Becoming-Smarter-than}Becoming Smarter than Albert
Einstein!}
\end{doublespace}

\begin{doublespace}
If our goal is to create artificial intelligence, (or anything else),
we should aim for the sky in the hope that we might at-least end up
reaching the treetops. This takes us to the central assumption of
this paper, which then becomes the ultimate benchmark to beat for
any intelligent system.
\end{doublespace}
\begin{assumption}
\begin{doublespace}
\label{assu:Albert-Einstein-is}Albert Einstein is the most intelligent
human being that has ever lived. It has been remarked, albeit anecdotally,
that his Intelligence Quotient (IQ) was between 160 to 190, give or
take a few points. For simplicity, and perhaps also because Albert
Einstein is more well know than other super smart individuals, we
overlook the fact that other people have recorded higher levels of
IQ.
\end{doublespace}
\end{assumption}
\begin{doublespace}
We wish to clarify that instead of Einstein, we could have used the
name of Nikola Tesla (or perhaps another remarkable individual) without
distorting the message from this section. We also completely stay
away from the debate about the limitations of using IQ as an indicator
of intelligence since it will not make a conceptual difference for
our discussion. (Weinberg 1989; Bartholomew 2004) describe the status
of controversies regarding the definition of intelligence, whether
intelligence exists and if it does whether it can be measured, and
the relative roles of genes versus environments in the development
of individual differences in intelligence. (Ceci \& Liker 1986) suggest
that IQ is unrelated to skilled performance at the racetrack and to
real-world forms of cognitive complexity that would appear to conform
to some of those that scientists regard as the hallmarks of intelligent
behavior. (DeDonno 2016) find that IQ fails to predict certain aspects
of learning of Hold'em poker, a game of skill with significant complexity
attributes resembling real-life activities such as stock market investing
and shopping for a home. (Okuda, Runco \& Berger 1991; Wagner \& Sternberg
1985; Sternberg 2018) discuss the importance not only of conventional
analytical intelligence but also skills needed for real world problem
solving such as common sense, creativity, knowledge that is usually
not expressed or taught, and wisdom that is not captured or hard to
measure using presently known standardized tests.

A few other interesting viewpoints are below. This includes intelligence
in man-made systems, which includes the possibility that our world
was created by some of us from the future or even the past after we
have evolved to transcend time. (Hernández-Orallo \& Dowe 2010) discuss
the idea of a universal anytime intelligence test, that is a test
that should be able to measure the intelligence of any biological
or artificial system that exists at this time or in the future. (Martínez-Plumed,
Ferri, Hernández-Orallo \& Ramírez-Quintana 2017) warn about the need
to be careful when applying human test problems to assess the abilities
and cognitive development of robots and other artificial cognitive
systems. (Hernández-Orallo, Martínez-Plumed, Schmid, Siebers \& Dowe
2016) contend that there is poor understanding about what intelligence
tests measure in machines and whether they are useful to evaluate
AI systems. They conclude that AI is still lacking general techniques
to deal with a variety of problems at the same time though a more
careful understanding of what intelligence tests offer for AI may
help build new bridges between psycho-metrics, cognitive science,
and AI. Though we believe that casting a wider net across all artificial
disciplines is necessary as discussed in Section (\ref{sec:The-Benchmark-for}).

Intelligence has to be a more multi-dimensional criteria. If we make
the simplifying assumption that we are somehow able to capture all
the attributes and higher dimensions of intelligence into a single
metric. That is, we are able to combine all the desirable features,
that help to deal with the uncertainty in our lives, for solving problems
into a single numeric score. We could call it IQ, which could still
be Intelligence Quotient or we could name it Infinite-Intellect Quest
or Imagination Quotient or better still, Involvement Quotient, for
lack of an even better term. (Kashyap 2018) tries to provide a more
complete measure of intelligence.

We could state that we live in a world that requires around 2000 IQ
points to consistently make correct decisions (Ismail 2014; End-note
\ref{enu:Taleb and Kahneman discuss Trial and Error / IQ Points}).
But the problem is that the best of us, by Assumption (\ref{assu:Albert-Einstein-is})
above, has less than 200 IQ points. Hence perhaps, to solve problems
flawlessly, we need someone like IQ-Man who might be friends with
Super-Man. For society's fascination with superheroes or super-humans,
see (Eco \& Chilton 1972; Reynolds 1992; Fingeroth 2004; Haslem, Ndalianis
\& Mackie 2007; Coogan 2009). But unfortunately, these supreme beings
are nowhere to be found. Super-Man at-least can be seen in movies.
IQ-Man is truly, as of now, nowhere to be found. He is not even there
in a comic book. Hence, the rest of us could use the clues mentioned
below, both for dealing with our problems and to create intelligence
in machines.
\end{doublespace}

We provide the below list of possibilities to address the question:
Can we become smarter than Albert Einstein?
\begin{enumerate}
\begin{doublespace}
\item \label{enu:The-Miraculous-Circle}The Miraculous Circle of Trial and
Error
\end{doublespace}
\begin{enumerate}
\begin{doublespace}
\item With each try and subsequent failure, we learn a way to improve and
move closer to success. Each improvement brings a better way to accomplish
something, or in a way enhanced IQ. But success lasts only till it
will fail and we need to try something else and start all over again
(Ismail 2014; End-note \ref{enu:Taleb and Kahneman discuss Trial and Error / IQ Points}).
Trial and error is easier said than done. Many people become adept
at a particular skill that develops because of repeated practice.
But it is much harder to obtain that in many real life situations.
(Kashyap 2018) suggests that if we develop the ability to spot similarities,
which is less natural compared to spotting differences since that
has more evolutionary backing, we might be able to apply what we learn
in more situations.
\end{doublespace}
\end{enumerate}
\begin{doublespace}
\item \label{enu:Lessons-from-other}Lessons from other Relevant Episodes
in History
\end{doublespace}
\begin{enumerate}
\begin{doublespace}
\item The errors need not all be due to our efforts. We can learn from instances
where similar things have been tried and see what we can glean from
the mistakes of others. For excellent introductions on the lessons
history holds, see: (Durant 1968; Malomo, Idowu \& Osuagwu 2006).
\end{doublespace}
\end{enumerate}
\begin{doublespace}
\item \label{enu:Team-Work}Team Work
\end{doublespace}
\begin{enumerate}
\begin{doublespace}
\item If a team of agents has the common purpose of accomplishing something
the effect is increased intelligence, as long as no one is looking
to sabotage the efforts of others. This is also known as the wisdom
of the crowd (Giles 2005). What one person might overlook another
might notice. The overall effect accomplished might be the betterment
of everyone involved.
\end{doublespace}
\end{enumerate}
\begin{doublespace}
\item \label{enu:Insatiable-Curiosity-and}Insatiable Curiosity and the
Desire to Learn
\end{doublespace}
\begin{enumerate}
\begin{doublespace}
\item Any agent that continues to be overwhelmingly curious, which will
lead to collecting new pieces of information, might continue to have
an uptick in the overall intelligence. (Reio Jr, etal 2006; Loewy
1998; Loewenstein 1994; Berlyne 1954; 1966; Litman \& Spielberger
2003) discuss the conceptualization and measurement of curiosity.
This suggests that our attempts at AI could have been misguided, what
we actually need to strive for can be termed artificial curiosity
and intelligence happens as a consequence of those efforts. But this
requires certain basic things to be established, which we will discuss
informally in Section (\ref{sec:A-Journey-to}) and more formally
in our road-map for intelligence in Section (\ref{sec:A-Road-map-for})
and provide the mathematical elements in Appendix (\ref{sec:From-Words-to-Symbols}).
\end{doublespace}
\end{enumerate}
\end{enumerate}
\begin{doublespace}
(Gopnik, Meltzoff \& Kuhl 1999) argue that evolution designed us to
both teach and learn. They indicate that nurture is our nature and
the drive to learn is our most important instinct. Perhaps as important
as, or even more important, than our instinct to survive. They reason
that even very young children, as well as adults, use some of the
same methods that scientists use to conduct research and to learn
about the world. (Campbell 1956) notes a formal parallel between some
of the characteristics of organic evolution and trial and error learning. 

Any discussion of children and grownups is incomplete without making
explicit when does childhood end? Here, we are not asking what is
childhood since that is perhaps harder to define. But it would be
a safe assumption that most humans have had somewhat of a childhood,
however brief that might have been.
\end{doublespace}
\begin{defn}
\begin{doublespace}
\textit{The end of childhood is when curiosity and confidence are
overtaken by the other concerns that life brings.}
\end{doublespace}
\end{defn}
\begin{doublespace}
New data show that infants use computational strategies to detect
the statistical and prosodic patterns in language input. This leads
to the discovery of phonemes and words (Kuhl 2004). (Oja 1982) derives
a new class of unconstrained learning rules using a simple linear
neuron model. He shows that the model neuron tends to extract the
principal component from a stationary input vector sequence. (McCulloch
\& Pitts 1943; Nass \& Cooper 1975; Takeuchi \& Amari 1979) are about
models on neuron activity and the many roles that have been assigned
to individual neurons from computational machines to analog signal
processors.

Language and learning is most likely to be a two way street. The rules
by which infants perceive information, the ways in which they learn
words, the social contexts in which language is communicated and the
need to remember the learned entities for a long time, probably influenced
the evolution of language (Kuhl 2004).

(Bush \& Mosteller 1955; 2006) present a mathematical model for simple
learning. Changes in the probability of occurrence of a response in
a small time are described with the aid of mathematical operators.
The parameters which appear in the operator equations are related
to experimental variables such as the amount of reward and work. (LeBlanc
\& Weber‐Russell 1996) present a computer simulation designed to capture
the working memory demands required in the comprehension of arithmetic
word problems, based on the belief that understanding arithmetic word
problems involves a complex interaction of text comprehension and
mathematical processes. 
\end{doublespace}
\begin{criterion}
\begin{doublespace}
\label{assu:To-learn-anything,-Language}To learn anything, any agent
first needs to learn a medium through which the learning can occur.
Simply put, to start to learn, we first need to learn a language.
\end{doublespace}
\end{criterion}
\begin{doublespace}
(Lenneberg 1967) hypothesized that language could be acquired only
within a critical period, extending from early infancy until puberty,
``the coming of language occurs at about the same age in every healthy
child throughout the world, strongly supporting the concept that genetically
determined processes of maturation, rather than environmental influences,
underlie capacity for speech and verbal understanding''. 

(Johnson \& Newport 1989) tried to check whether it should be the
case that young children are better second language learners than
adults and should consequently reach higher levels of final proficiency
in the second language. They tested the English proficiency attained
by 46 native Korean or Chinese speakers, who had arrived in the United
States between the ages of 3 and 39, and who had lived in the United
States between 3 and 26 years by the time of testing. Their study
supported the conclusion that a critical period for language acquisition
extends its effects to second language acquisition. (Newport 1990)
considers evidence from several studies of both first and second language
acquisition suggesting that normal language learning occurs only when
exposure to the language begins early in life.

(Sutton \& Barto 1998) provide an excellent introduction to understand
intuitively the ideas of reinforcement learning and the general connection
between its parts. They define reinforcement learning as learning
what to do and how to map situations to actions, so as to maximize
a numerical reward signal.

It is interesting to note that there is contrasting evidence. (Snow
\& Hoefnagel-Höhle 1978) test the hypothesis that second language
acquisition will be relatively fast, successful, and qualitatively
similar to first language only if it occurs before the age of puberty.
They studied the naturalistic acquisition of Dutch by English speakers
of different ages. It was found that the subjects in the age groups
12-15 and adults made the fastest progress during the first few months
of learning Dutch and that at the end of the first year, (the subjects
were tested 3 times during their first year in Holland to assess several
aspects of their second language ability), the 8-10 and 12-15-year-olds
had achieved the best control of Dutch. The 3-5 year-olds scored lowest
on all the tests employed. These data do not support the critical
period hypothesis for language acquisition. This perhaps suggests
that we need good command over one language before we can learn another
language.

A point we need to keep in mind is that perhaps, English language
(and many languages), especially its pronunciations and grammar, is
not the easiest to learn due to the many nuances it has that do not
generalize easily. It would then make sense to develop a language
that is more structured and free of ambiguity. (Stageberg 1968) has
a discussion of structural ambiguity with some examples. It is important
to note that this is not just about designing a language with very
precise syntax. It is equally important to convey the semantics and
the context of situations in greater detail. With human interactions
we are able to assume this understanding of the context in many instance,
though it creates some confusion in some places (End-note \ref{enu:Mother-Daughter}).
But with artificial agents it might be necessary to provide greater
clarity on the situations and supply more detailed contextual backgrounds.
\end{doublespace}
\begin{doublespace}

\section{\label{sec:A-Journey-to}A Journey to the Land of Unintended Consequences}
\end{doublespace}

\begin{doublespace}
A glimpse of what a journey towards the land of unintended consequences
holds can be seen by reminding ourselves that all knowledge creation
is but an unintended consequence. We start with an attempt to understand
the papers written by others and end up with papers of our own. That
is, beginning with the literature review of knowledge already created
or trying to understand experiments performed and under what conditions,
we arrive at results that add what is missing or suggest improvements.
Although to be precise, as researchers, we do want to intentionally
create new knowledge, but the exact new knowledge we end up creating
is unintentional. We stumble upon new knowledge as we wander around
the knowledge that is already in place. This is simply because our
intentions can only cater to what we already know, or, to existing
knowledge. New knowledge, which is unknown, has to come from the realm
of the unintentional.
\end{doublespace}
\begin{doublespace}

\subsection{\label{subsec:I-Don't-Know,}I Don't Know, A Great Answer}
\end{doublespace}

\begin{doublespace}
(Taleb 2007) in his landmark book, the Black Swan, talks about the
unread books in the personal library of legendary Italian writer,
Umberto Eco, and how over time this unread collection gets larger.
Hence, it would not be incorrect to say that there is more that we
don’t know than what we know. The more we know, the more there will
be to know. But that should not stop us, and the agent, from trying
to seek the answers or even from making a guess as a starting point.

Hence, an answer admitting ``I Don't Know'' is a great answer in
most situations. When we design any system or model, especially in
AI, questions and answers are important since that is the primary
way to assess the presence of intelligence. But what becomes more
important are our definitions and assumptions. To supplement our definition
of intelligence (Definition \ref{def:Intelligence-is-the}) we provide
the following cardinal assumption.
\end{doublespace}
\begin{assumption}
\begin{doublespace}
\label{assu:The-knowledge-that}The knowledge that has been accumulated
over time is lesser than the knowledge that is yet to gathered. With
this assumption, an answer of ``I Don't Know'' becomes not just
a correct answer, but it is an invitation to the person asking the
question to teach the agent how to answer the question.
\end{doublespace}
\end{assumption}
\begin{doublespace}
So the agent is always learning and the reason is simply due to what
we discussed before. We don’t know most things and hence the learning
usually never stops. If the person asking the question is not satisfied
with the answer, he or she now has a responsibility to teach the agent
to improve upon the answer produced. A failure to create intelligence
in any agent is a failure on the part of the teacher in finding a
teaching methodology appropriate for the agent. This also implies
the next criteria.
\end{doublespace}
\begin{criterion}
\begin{doublespace}
\label{cri:Creating-intelligence-is}Creating intelligence is not
only about writing software code, it is about having the best teachers
that humanity has produced being available to teach the later generations,
be it human or machines.
\end{doublespace}
\end{criterion}
\begin{doublespace}
We now consider the fundamental question of whether we need complicated
models or merely stronger beliefs. We state this as our essential
doctrine.
\end{doublespace}
\begin{criterion}
\begin{doublespace}
\label{cri:The-agent-has-Self-Confidence}The intelligent agent has
to believe that it has the ability to learn and the confidence to
request lessons regarding answers that it is unable to generate satisfactorily.
\end{doublespace}
\end{criterion}
\begin{doublespace}
Confidence, like intelligence, is an unintended consequence. We cannot
find confidence directly or our actions cannot become confident just
by our choice to do so. As an illustration, let us say someone has
bad vision and they decide to walk around confidently. They might
not only cause harm to themselves but they are a disaster for everyone
around them. To build confidence we need to seek clarity or we need
to focus our efforts on seeing things clearly. Once there is better
sight, it will lead to a more confident walk. An admission of ignorance
regarding something or acceptance that we don't know becomes a great
possibility to know. This opportunity marks the start of gaining confidence.

As an unintended consequence of our struggle to try and comprehend
things around us better, we gradually become confident as our understanding
improves. Combining confidence, or our pains to pursue clarity, with
the great answer ``I don't know'', which follows from Assumption
(\ref{assu:The-knowledge-that}), we get a better answer which is
``let me try''. In addition, it is worth highlighting that the discussion
surrounding Assumptions (\ref{assu:Albert-Einstein-is}; \ref{assu:The-knowledge-that}),
about our limited intelligence in comparison to the intelligence of
the creation around us which causes the uncertainties that we experience,
should help to keep overconfidence in check.

When an agent is not learning it should ideally be teaching (other
agents or anyone else). This is because teaching and learning are
highly interconnected and the best way to learn is to teach. A realization
that the roles of students and teachers are constantly getting interchanged
originates from a belief that everyone has something to teach to everyone
else. When we are teaching we are also learning from someone else.
When we are learning we are really teaching ourselves. To be clear,
although most of us probably know this, learning does not just represent
reading textbooks or doing assignments, though these are important
components of learning. Learning can happen when we are doing anything
that we enjoy doing. This can be built into the reward system of the
agent so that it accumulates points for aspects that it likes. Different
agents could be made to like different things so that we build a random
enjoyment component that learns from different activities.

Hence, if any agent has to learn a lot (or everything really?), instead
of trying to find the right teachers we should make everyone its teacher.
Since we have to respect our teachers, the agent now has to respect
everyone. A consequence of everyone becoming a teacher, and since
the roles of teacher and student can interchange, is that everyone
also becomes everyone else's student. And the result might be that
everyone will respect one another. Isn't that one of the objectives,
and perhaps an unintended consequence, of making everyone intelligent?

Many times what we don't know, or even when we are in a situation
where we don't know something, can be scary or can cause confusion
or frustration. Hence, efforts at learning and teaching usually end
up confronting these two monsters: Confusion and Frustration. Both
of these, though scary and ugly to begin with, can be powerful motivators
as long as we don’t let them bother us. Confusion is the beginning
of Understanding. Necessity is the mother of all creation / innovation
/ invention, but the often forgotten father is Frustration, which
is sometimes even more necessary than necessity herself. Simply put,
some amount of frustration can be highly stimulating and lead to great
possibilities. What we learn from the story of, Beauty and the Beast,
(De Beaumont 1804; End-note \ref{enu:Beauty-Beast}), is that we need
to love the beasts to find beauty. Hence, if we start to love these
monsters (Confusion and Frustration), we can unlock their awesomeness
and find truly stunning solutions.

Hence, our agent has to remain confident and ask questions when it
does not have an answer. This can be stated as,
\end{doublespace}
\begin{sol}
\begin{doublespace}
\textit{\label{cond:Life-for-an}Life for an intelligent agent is
all about having confidence and the right teachers and / or students.}
\end{doublespace}
\end{sol}
\begin{doublespace}

\subsection{\label{subsec:Acing-the-Turing}Acing the Turing Test }
\end{doublespace}

\begin{doublespace}
(Moor 1976) puts forth the argument that the real value of the imitation
game (also known as the Turing Test, TT, Turing 1950; End-note \ref{enu:The-Turing-test})
lies not in treating it as the basis for an operational definition,
but in considering it as a potential source of good inductive evidence
for the hypothesis that machines think. (French 1990) argues that
the very capacity of the TT to probe the deepest, most essential areas
of human cognition makes it virtually useless as a real test for intelligence.
(French 2000) chronicles the comments and controversy surrounding
the first fifty years of the TT. He concludes that it will remain
important and relevant to future generations of people living in a
world in which the cognitive capacities of machines will be vastly
greater than they are now.

(Copeland 2000) suggests, based on unpublished material by Turing,
that the Turing test withstands objections that are popularly believed
to be fatal. (Harnad 1992) shows that it is important to understand
that the TT is not, nor was it intended to be, a trick. How well one
can fool someone is not a measure of scientific progress. The TT is
an empirical criterion. It sets AI's empirical goal to be to generate
human scale performance capacity. This goal will be met when the candidate's
performance is totally indistinguishable from a human's. Until then,
the TT simply represents what it is that AI must endeavor eventually
to accomplish scientifically.

(Saygin, Cicekli \& Akman 2000) conclude that the Turing Test has
been, and will continue to be, an influential and controversial topic.
(Von Ahn, Blum \& Langford 2004) discuss the Completely Automated
Public Turing Test to Tell Computers and Humans Apart, CAPTCHA. This
is a automatically generated test, which most humans can pass but
that current computer programs cannot pass. This is somewhat of a
paradox since a CAPTCHA is a program that can generate and grade tests
that it itself cannot pass. This finds application in many places
on the internet to ensure that computer programs are not substituting
for humans.

An often omitted criteria that needs to be considered when administering
the Turing test is the ability, or, the level of skill of the person
conducting the test. Surely, different individuals are satisfied with
different levels of impersonation. When we see any movie (play or
drama) that depicts the life of any real person, while reminding ourselves
that movies might not be real but real life can become movies, different
people are satisfied with different levels of acting ability. We all
know that the person playing the role in the theatrical version is
not the same individual as the person that is being enacted. But in
many cases, (perhaps, in most cases when it is well produced), we
leave feeling satisfied with the result of the replication. The lesson
for us here is this. How far does the test administrator need to go
to believe that the computer program perfectly duplicates the human
test subject?

Our Definition (\ref{def:Intelligence-is-the}) of intelligence implies
that the benchmark for intelligence has been surpassed if the question
is answered to the satisfaction of the person administering the Turing
test. In this context answering a question is the decision making
on display.
\end{doublespace}
\begin{doublespace}

\subsection{\label{subsec:Imitation-in-the}Imitation in the Imitation Game}
\end{doublespace}

\begin{doublespace}
Let us now consider another example of imitation in the imitation
game, which was a recently released movie about the role of Alan Turing
in the second world war (Proudfoot 2015; You 2015; Guo 2015). The
actor in the movie, Benedict Cumberbatch (Porter 2014), does a sensational
job portraying the real Alan Turing. Though this is a subjective evaluation,
if someone disagrees, termed a disbeliever, then it would be fair
to state that they now have the responsibility of doing a better role
play. To go into length on how Benedict Cumberbatch (or any disbeliever,
forced to turn into a better actor) accomplished this would require
another paper or a few books of their own. (Hagen 1991; 1973) are
masterpieces on how to be convincing actors. The short answer would
be that an actor believes that he can play the part he is chosen to
play. This is what an agent chosen to display intelligence must first
be made to believe. This is about not about dishonesty or deception,
it is about belief and confidence. As discussed in Section (\ref{subsec:I-Don't-Know,}),
true confidence comes when we admit we don't know something and we
are willing to try.

The manner in which Benedict Cumberbatch (End-note \ref{enu:Certain-six-year})
plays the main character in the movie, Imitation Game, leads us to
state the Real Enigma of the Imitation Game. Which Alan Turing is
the More Convincingly Brilliant Mathematician? This question merely
inquires as to whether, Alan Turing or Benedict Cumberbatch, would
pass a stage test for actors who had to convince the audience they
were mathematicians. Anyone that would make the argument that acting
like a mathematician does not make a real mathematician, needs to
be reminded that acting like a mathematician is the first step to
being a mathematician (End-note \ref{enu:Infinite-Progress}). Once
this belief is instilled time and familiarity with the steps and notation
related to mathematics, supplemented with our road-map for intelligence
(Section \ref{sec:A-Road-map-for}), will take care of creating real
mathematicians. (Kashyap 2017) is an application of our curious and
confident approach to creating intelligence in the financial markets.
\end{doublespace}
\begin{doublespace}

\subsection{\label{subsec:Mexican-Chihuahua-solving}Mexican Chihuahua solving
Korean Puzzles under a Mush-Room}
\end{doublespace}

\begin{doublespace}
(Searle 1980) argued that the fact that machines can be devised to
respond to input, with the same output that a mind would give, does
not mean that minds and machines are doing the same thing, for the
latter lacks understanding. (End-note \ref{enu:Chinese room thought experiment})
has a summary of Searle's Chinese-Room thought experiment. (Searle
1982; 1990; 2001; 2004) are later discussions. (Preston \& Bishop
2002) has a collection of essays on this crucial challenge. Searle
was in fact against the notion of strong AI, which is that human minds
are in essence computer programs. That is, an appropriately programmed
computer with the right inputs and outputs would thereby have a mind
in exactly the same sense human beings have minds. All mental activity
is simply the carrying out of some well-defined sequence of operations
frequently referred to as an algorithm.

(Penrose 1989) claims that there are aspects of consciousness that
cannot be replicated within any computer model, no matter how sophisticated,
as long as the model is based on an algorithm. He presents an overview
of the present state of physical understanding and tries to show that
an important gap exists at the point where quantum and classical physics
meet. He speculates on how the conscious brain might be taking advantage
of whatever new physics is needed to fill this gap to achieve its
non-algorithmic effects. 

Searle's example has had a profound impact on the discussions related
to AI for the last many years. However, as a counterargument, we pose
this alternate scenario. Instead of an American (John Searle), juggling
with Chinese characters he has no clue about, in a closed room using
instructions in English a language he understands. Let us consider
a Mexican Chihuahua solving puzzles posed using Korean characters,
seated under a giant Mush-Room. Perhaps having devoured some of the
mushroom, the Chihuahua is being influenced by it in ways that we
do not yet quite comprehend. But for the purposes of this test the
effects are only beneficial. For the hallucinogenic effects of mushrooms
see: (Schwartz \& Smith 1988; Samorini 1992; Musshoff, Madea \& Beike
2000; Halpern 2004). The Chihuahua is giving out the right answers
to the puzzles back in the form of Korean characters, but only barks
in response to everything else.

Does it matter whether the Chihuahua is only using certain training
it has been given to use rules to arrange Korean characters? Or, whether
it is the Mushroom causing the miracle or something else? For all
practical purposes, the Chihuahua is an intelligent creature since
it is able to present the right set of Korean characters as a solution
to the puzzles or questions we pose. The Chihuahua simply does not
speak the same language as we do. We do not understand its barking
nor does it understand the voice tones we produce. Or maybe, it pretends
that it does not understand what we say. It can be argued, though
we won't continue this line of reasoning, that we understand less
of what dogs say than what dogs understand of what we say. Who is
more intelligent then? For simplicity and for rhetorical reasons,
let us just say that the effects of the mushroom last for as long
as the Chihuahua is alive, or, until we are still interested in asking
it questions using Korean Characters?
\end{doublespace}
\begin{summary}
\begin{doublespace}
Let us substantiate this counter viewpoint. We completely believe
that we understand the solution and we rely on rules to arrive at
the solution. Does it really matter if we are simply using rules to
solve a puzzle or if we are actually understanding how the solution
was arrived at? This is not about being dishonest, or, passing lie
detector tests. If we believe we know the answer and if we are able
to consistently generate the answer, it does not matter how we got
the answer. Understanding then becomes a state of mind or a belief.
We should now be deemed intelligent enough as we have come up with
the answers.
\end{doublespace}
\end{summary}
\begin{doublespace}

\subsection{\label{subsec:Merry-Go-Round-of-Decisions,}Merry-Go-Round of Trials,
Errors and Revisions}
\end{doublespace}

\begin{doublespace}
Usually, on our first attempt to answer any question we may not get
the correct or the best answer. This is where the trial and error
part kicks in. But once we start somewhere, we learn from our mistakes
and improve upon our explanations. In this Question \& Answer context,
we define any question as a good question and a good answer as something
that we only think of later. That is, a good solution is something
we find after a few iterations of trial and error.

(McCarthy \& Hayes 1969) is a discussion of the main issues in philosophy
that also arise in AI. John MaCarthy, who is credited with coining
the term “Artificial Intelligence” defined it as “the science and
engineering of making intelligent machines” (McCarthy 2004). (Beck
\& Arnold 1977) discuss this iterative approach to estimate parameters
used in Engineering and the Sciences. Many improvements in the sciences
and engineering happen through a series of refinements.

(Wolfe 2005) is a discussion of how successive designs of fighter
planes, where a failure potentially meant the loss of life of the
pilots, brought us incremental improvements and eventually made the
possibility of space exploration a reality. End-note (\ref{enu:Taleb and Kahneman discuss Trial and Error / IQ Points})
is a mention by Taleb of why it is important to create an environment
where the errors are less costly, or, why trials with small errors
are preferable. Though sometimes expensive errors are unavoidable
as in plane crashes, which subsequently led to safer air travel for
later passengers. (Phillips, Lin, Schifter \& Folse 2019) suggest
the adoption of a piecemeal engineering or tinkering approach, augmented
by adaptive policies and modern collaboration platforms, to maximize
the prospects of sustainable practices worldwide.

(Swanson 1977) recognizes the essential role of trial and error in
accessing to scientific literature. This points the way toward improved
information services illuminating potential inconsistencies that have
beset many retrieval exercises. This has strong implications for our
knowledge store discussed in Section (\ref{subsec:Knowledge-Store}).

(Doidge 2007) presents classic cases from the frontiers of neuroscience
that chronicle the biological changes happening in the brain driven
by external impetuses. This reveals that adapting to new circumstances
and learning to deal with adversity are almost hard wired into us.
In essence what this discloses is that the brain constantly changes
as situations change, which tells us that what we need to contend
with or mimic, in our AI ambitions, is a moving target.

(Young 2009) is about trial and learning in a social or economic game
theory setting (Gibbons 1992). A person learns by trial and error
if he occasionally tries out new strategies, rejecting choices that
are erroneous in the sense that they do not lead to higher payoffs.
In an economic game, however, strategies can become erroneous due
to a change of behavior by someone else triggering a search for new
and better strategies. In economics, it is insightful to establish
conditions under which the Nash equilibrium property (Nash 1950) can
be established. But in real life equilibrium is a dynamic, constantly
changing state (like a see-saw) due to the subjectivity in all decision
making and the differing perceptions of the individuals involved.
Hence the trial and error never ceases.

Intelligence and learning also involve the ability to guess or the
ability to make decisions when the best choice is not exactly clear.
Observed data can be consistent with many models, and therefore which
model is appropriate, given the data, is uncertain (Ghahramani 2015).
Similarly, predictions about future data and the future consequences
of actions are uncertain. Probability theory provides a framework
for modeling uncertainty. A machine can use such models to make predictions
about future data and take decisions that are rational given these
predictions.

In all efforts at creating intelligence, we make an unstated assumption
that human beings are capable of intelligence. But, we are not born
intelligent. It takes years of nurturing and tutoring for us to become
intelligent. We display different abilities and aptitude for different
things, or the intelligence of different individual could be in different
skills. How could we then have expectations that something, that we
deem not to have the capacity for intelligence, has to become intelligent
in a relatively short span of time? This holds a strong message for
us that to create intelligence artificially might require years of
training for an agent.

In a typical classroom some kids end up doing better, in terms of
conventional forms of intelligence in comparison to others, as assessed
by our benchmarks or measures. This is due to the creation of more
connections and better retention of the relevant bits of information
they receive. Using our Assumption (\ref{assu:Albert-Einstein-is}),
we can reword this as follows. In a world full of intelligent human
beings, only a handful of us become Albert Einsteins. Hence, we could
expect a similar sort of situation when trying to create AI. We need
to start with a group of agents, with different parameters, and let
them wander around and see what innate abilities they pick up. Accordingly,
we need to further those skills that were naturally (or probabilistically)
acquired. The circle of trial, error and corrections needs to be happening
constantly.
\end{doublespace}
\begin{doublespace}

\subsection{\label{subsec:Gifts-for-AI}Gifts from the Realm of the Unintentional}
\end{doublespace}

\begin{doublespace}
(Fogel 2004) chronicles that infantile amnesia, the apparent loss
of memory about one's own infancy, has been accepted as fact for at
least a few thousand years. (Waldfogel 1948) reveals a serious gap
in our knowledge regarding childhood memories. This is despite the
abundance of clinical evidence regarding the fact that repressed childhood
experiences may be significant for adult behavior. The evidence, though
plenty, cannot be constructed as proof of the universality or the
predominance of infantile amnesia due to the authenticity of the data
used in these studies.

(Nadel \& Zola-Morgan 1984) indicate that some memory systems in our
body become functional at birth or shortly thereafter, whereas others
become active following a period of postnatal neurogenesis. Also,
studies have shown that localized brain damage typically leads to
selective rather than general memory defects. This suggests that the
postnatal maturation of a specific neural system lies at the root
of infantile amnesia. (Howe \& Courage 1993) conclude that infantile
amnesia is a chimera of a previously unexplored relationship between
the development of a cognitive sense of self and the personalization
of event memory. They examine this hypothesis in the context of related
developments in language and social cognition. (De Brigard 2014) is
a philosophical discussion of the phenomenon of remembering along
with a historical perspective including reviews of critical findings
in the psychology and the neuroscience of remembering.

Despite the many unknown aspects of infantile amnesia, it is clear
that the formative years of any human being are not remembered. Perhaps,
an unintended consequence of not knowing who we really were, before
we got a better idea of who we were becoming, is to reduce any anguish
as we learn to explore and form a conception of what we are. Or maybe,
evolution only deemed worthy of remembering only what we remember,
which is after we had a better idea of what was happening around us.

In addition, perhaps the most important element of AI is to ignite
curiosity within the agents. Because, once an agent gets inquisitive,
learning happens almost by itself after that. An unintended consequence
can be overconfidence and needs to be monitored for closely (Section
\ref{subsec:I-Don't-Know,}). At periodic intervals, the agent has
to be corrected so that positive learning is rewarded and mistakes
are reversed. To prevent the abuse of excessive intelligence, perhaps,
the teachers who train the agent also need to impart moral behavior
and empathy towards the, so called, less intelligent.

To triumph in creating intelligence, and almost everything else, it
is important to know where we are and start the journey towards where
we want to be. Sometimes that might mean a change in direction. And
changing direction, even slightly, could be defined as the start of
a new journey. A consequence (perhaps unintended) of taking the first
step on a journey means that the percentage progress we have made,
in terms of the distance travelled, shoots up to infinity (End-note
\ref{enu:Infinite-Progress}). So once we start the trip it becomes
manageable immediately. The subjectivity in how we compare things
means that the benchmark for AI will be constantly changing. This
means we need our agents to keep on learning just as we need to do
the same as well.

A further glance in the direction of unintended consequences might
show that in the process of creating knowledge or intelligence, and
trying to understand the world better or make it a better place, we
might just end up understanding one another better. Perhaps, becoming
more tolerant in the process, an unintended yet very welcome consequence.
This should make us wonder whether the the true purpose of all knowledge
or intelligence creation might be to make us more tolerant.

As AI becomes tightly interwoven with many aspects of our daily lives,
another unintended consequence would be the many jobs that would no
longer need any human intervention. While on the surface this might
seem like a grave threat. This trend would force human beings to look
inward into what truly makes them human and realize the greater potential
of their minds. This also highlights the key strength that we possess.
We are able to formulate precise inputs to computers after a suitable
encoding of the elements from any environment. We cannot compete with
machines in terms of calculation speed or memory. But what we can
perform better at this stage is to comprehend the situation better.
This suggests that our advantage is being able to figure out what
the real problem or challenge is.
\end{doublespace}
\begin{doublespace}

\subsection{\label{subsec:Evolutionary-Tricks-for}Evolutionary Tricks for the
Empirics}
\end{doublespace}

\begin{doublespace}
Anyone reading this paper might have certain well founded reservations,
regarding the theoretical and conceptual connections outlined here,
because of the apparent lack of empirical tests. For that, we would
like to point out that the present paper is based on millions of years
of experimentation (Brooks 1991). This test, which is still going
on and which is still creating intelligence, is nothing but evolution
(Darwin 1859; Dawkins 1976; Eldredge 2005; Scott 2009). To replicate
this test might take another few millions years. But perhaps, it can
be done using machines in a shorter time, though it might still take
a few years. This paper also provides the theoretical basis for many
new empirical tests and is based on one (or many?) continuous and
complete empirical experiment(s) happening everywhere in our cosmos.

As discussed in Section (\ref{subsec:Merry-Go-Round-of-Decisions,}),
this cycle of revisions is also at play in the numerous species that
inhabit our planet. A large avenue for future research would be to
explore the level of intelligence in different kinds of creatures
and the extent to which they display the characteristics described
here. It is important to be open to other fundamental principles omitted
here, since intelligence is also about being open to possibilities.
This might also reveal that evolution and its twin sister, reproduction,
are passing on genetic improvements geared towards increased intelligence
to later generations. Such studies rooted in the basics of biology
might even reveal bottlenecks to increasing intelligence based on
the physical properties of any system. We just need to keep reminding
ourselves that what happens is usually unintentional, but it just
might bring about wonderful consequences once it happens. Remembering
this matters since some (most?) of us might only see part of the benefits,
or, the impediments, at any point in time.
\end{doublespace}

When it comes to intelligence there might be an overemphasis on the
brain, even though every cell in our body has remarkable intelligence
about what it is supposed to do (Kashyap 2018b). Evolutionary theory
suggests that life started from a single celled organism that became
a human being with a complex brain. So a single cell is capable of
creating a complex brain given enough time and opportunities to try
and learn, also known as, evolution with trial and error. We have
to consider the possibility that there is intelligence inside every
single cell in every single organism that can or cannot be seen. 

The empirical lessons from this intelligence, present all around us,
are manifold. We can perform tests to detect varying levels of decision
making in different settings around us. The decision goals in these
situations could be limited, depending on what is chosen for evaluation,
but could hold valuable lessons. What we discern from these experiments
could be tailored for specific uses in artificial systems. It is highly
plausible that we are merely seeing different manifestations of this
omnipresent intelligence by the various aspects we are able to sense
using our sense organs, measurement devices and the brain. Our ability
to sense or perceive this intelligence everywhere might be limited.
Just because we cannot see something, does not mean it is not there.
As our investigative devices improve, we can hope to have access to
an increasing source of highly efficient techniques for decision making.
\begin{doublespace}

\section{\label{sec:A-Road-map-for}A Road-Map for Intelligence}
\end{doublespace}

\begin{doublespace}
A detailed axiomatic approach to uncertainty, unintended consequences
and sapience is postponed for another time, or perhaps, another lifetime.
The present assortment can be summarized as the below ``how to guide
for intellect'', or, a road-map of the essential elements required
to create artificial intelligence. The related concepts are elaborated
in Sections (\ref{subsec:I-Don't-Know,}; \ref{subsec:Acing-the-Turing};
\ref{subsec:Mexican-Chihuahua-solving}; \ref{subsec:Merry-Go-Round-of-Decisions,};
\ref{subsec:Gifts-for-AI}) and linked to the corresponding items
below.

\textit{\textcolor{black}{Each step in the following algorithm or
pseudo-code can be tested as a separate scientific hypothesis. But
surely, greater the coherence between the components that encapsulate
the below concepts better the intended outcomes. Relevant evidence
and technical aspects, including pointers to mathematical ingredients
from Section (\ref{sec:From-Words-to-Symbols}) and further references,
are given in the corresponding points below.}}

\textbf{\large{}While (Agent is Alive or The World has not Ended)}{\large\par}

\textbf{\large{}Begin}{\large\par}
\end{doublespace}
\begin{enumerate}
\begin{doublespace}
\item \label{enu:A-language-certification}A language certification is necessary.
\end{doublespace}
\begin{enumerate}
\begin{doublespace}
\item From Criterion (\ref{assu:To-learn-anything,-Language}), we need
to ensure that the agent can pick up advanced concepts by having been
certified previously that a certain minimum level of language abilities
have been acquired. If a certain threshold is not met in terms of
language skills, it is back to the language classroom for this agent
(Solution \ref{cond:Life-for-an}). (Lightbown, etal 1999) is a comprehensive
discussion on how languages are learned, especially from the point
of interest of classroom teachers.
\item To be clear, the requirement from the agent can be something simple
like giving advice on a financial strategy. In this case, the inputs
can simply be the time series of numbers. The output can be just a
Buy, Sell or Hold indication since all of finance through time has
involved only these three simple decisions. The complications are
mostly to get to these three outcomes, which the agent can conjure
up in its own way. But its interface with the external world need
not be anything too involved.
\item On the other extreme, if we wish to create agents that are mathematicians,
we would need a precise representation of mathematical rules and results
using explicitly clear notation and terminology. Both the input and
output have to adhere to exactly defined and rigorous statements.
The language in this case becomes a sequence of steps that flow from
the preceding one to the subsequent one using the rules of mathematics
or already established results or based on logical arguments. To begin
with, we only need to capture basic operations in mathematics and
more advanced results will be connected to the basic results using
fundamental rules or results already established. As discussed in
Section (\ref{sec:The-Benchmark-for}), breeding true intelligence
requires us to transcend artificial domains created by us, such as
finance or mathematics. But to make the problem of implementation
more manageable, focusing on simplified language requirements specific
to any field is prudent.
\end{doublespace}
\end{enumerate}
\begin{doublespace}
\item \label{enu:A-formal-model-collector}A formal model that collects
new pieces of information from the various possible choices.
\end{doublespace}
\begin{enumerate}
\begin{doublespace}
\item We model collection of information using the Bass Model of Diffusion
(Bass 1969; Mahajan 1985; Mahajan, Muller \& Bass 1991; Bass, Krishnan
\& Jain 1994; Michalakelis, Varoutas \& Sphicopoulos 2010; Jiang \&
Jain 2012; End-notes \ref{enu:Bass-Model-Diffusion}; Section \ref{subsec:Bass-Model-of}).
This is used extensively in marketing to study the adoption of new
products. We view a new product being adopted by someone as being
equivalent to the agent collecting the adopter. The person that just
newly adopted the product is the new piece of information from the
perspective of our model of information collection. So there is a
certain element of randomness in terms of which information comes
next. Collecting new pieces of information is how we mimic curiosity
in our agents.
\item New models of curiosity would benefit immensely from suggestions on
how to develop children to be lifelong learners and to nurture the
scientist within all kids (Calkins \& Bellino 1997; Ramey-Gassert
1997; Gamble \& Cota-Robles 2015; End-note \ref{enu:The-more-curious-baby}).
Here, we illustrate with the two opposite ends of the spectrum mentioned
in Step (\ref{enu:A-language-certification}). The new piece of information
could be a new time series of financial data if our requirement is
related to finance, based on the expectation that the agent has to
give advice on a financial strategy. The new piece of information
could be a new mathematical result if our requirement is mathematical,
based on the expectation that we wish to create mathematicians.
\end{doublespace}
\end{enumerate}
\begin{doublespace}
\item A collection of agents that pick different pieces of information using
the collector model.
\end{doublespace}
\begin{enumerate}
\begin{doublespace}
\item This is simply a group of agents with different parameters of innovation
and adoption for the Bass Model of diffusion (Section \ref{subsec:Bass-Model-of}).
We could also use many developments in the use of models of curiosity.
(Sato, Takeuchi \& Okude 2011) present an experience-based curiosity
model which indicates individual’s real time curiosity within a city
regarding how well the individual knows the city. It aims to understand
individual’s real time interests by not relying on information the
people input intentionally, but by understanding behavior data. This
is done through environmental sensing from mobile devices that are
capable of sensing the environment around the individual and not necessarily
by interacting with the people directly.
\end{doublespace}
\end{enumerate}
\begin{doublespace}
\item \label{enu:A-belief-in-Self}A belief instilled in the agent that
it/he/she is intelligent.
\end{doublespace}
\begin{enumerate}
\begin{doublespace}
\item This has been captured in Criterion (\ref{cri:The-agent-has-Self-Confidence})
and Solution (\ref{cond:Life-for-an}). To elaborate, confidence in
one’s abilities generally enhances motivation making it a valuable
asset for individuals with imperfect willpower (Bénabou \& Tirole
2002). For this the agent has to be an actor, believing that it can
play the part of someone who is intelligent. There is a vast literature
on the importance of self-confidence and its relation to performance
in different fields such as sports and language acquisition: (Feltz
1988; Clément, Dörnyei, \& Noels 1994; Noels, Pon, \& Clément 1996).
An overemphasis on global self-esteem is perhaps not ideal. (Owens
1993) discusses the implications for understanding the differential
impact of negative and positive self-evaluations on emotional and
social well-being.
\end{doublespace}
\end{enumerate}
\begin{doublespace}
\item \label{enu:A-measure-to}A measure to judge how closely the new information
collected matches the information already stored.
\end{doublespace}
\begin{enumerate}
\begin{doublespace}
\item To aid in this effort to extract meaning from chaos, we summarize
the application of the theoretical results from (Kashyap 2019) to
AI studies. The central concept rests on a novel methodology based
on the marriage between the Bhattacharyya distance, a measure of similarity
across distributions, and the Johnson Lindenstrauss Lemma, a technique
for dimension reduction. This combination provides us with a simple
yet powerful tool that allows comparisons between data-sets representing
any two distributions. Perhaps, also becoming to our limited knowledge,
an example of perfect matrimony (Sections \ref{subsec:Bhattacharyya-Distance};
\ref{subsec:Dimension-Reduction}). This methodology is necessary
to assess how similar newly gathered information is to the knowledge
store that we already have (Section \ref{subsec:Knowledge-Store}).
\item A subtle point here is that we might keep receiving, or collecting,
the same information multiple times. The degree of similarity of the
newly received information with the information we have already collected,
condensed (dimension reduced) and stored will increase over subsequent
iterations. Hence after a few rounds of certain information being
repeatedly received, it will be stored almost in its entirety. This
captures the fundamental principle of how learning happens by repetition.
The specific implementation models will need to fine tune the number
of times something is received before most of it is retained. But
at a high level this crucial concept has to happen. This also tells
us that the more often some information is being received, the more
important it is and the more completely it needs to be stored. The
case where we need a lot of precision is for any new mathematical
result, which has to be an exact combination of fundamental rules
or already established results.
\end{doublespace}
\item (Kashyap 2019) suggests that the Bhattacharyya distance might have
some advantages compared to other measures, such as the Kullback-Leibler
(KL) divergence. But there might be some instances where the KL divergence
or other distance measures might be better suited. Hence, being open
to different ways of comparing similarity is sensible.
\end{enumerate}
\begin{doublespace}
\item \label{enu:A-method-to-reduce}A method to keep reducing the information
store so that less data, the most essential (only almost a summary),
needs to be maintained.
\end{doublespace}
\begin{enumerate}
\begin{doublespace}
\item Any existing compression technique can be used (Lynch 1985; Storer
1988). (Shlens 2014) is about Principal Component Analysis for dimension
reduction, which utilizes variance maximization methods. The method
from Step (\ref{enu:A-measure-to}), can be used as well to reduce
the data store to a smaller dimension.
\end{doublespace}
\end{enumerate}
\begin{doublespace}
\item A regular period of deep sleep.
\end{doublespace}
\begin{enumerate}
\begin{doublespace}
\item The importance of sleep in human beings in not fully understood, but
it is beyond debate that sleep is essential and has numerous benefits
(Robertson, Pascual-Leone \& Press 2004; Ellenbogen 2005; Blischke,
etal 2008; Aly \& Moscovitch 2010; Nere, etal 2013). Any agent requires
a period of sleep, where the Steps (\ref{enu:A-measure-to}; \ref{enu:A-method-to-reduce})
are carried out without any other external disturbances. This is to
ensure that its confidence is not shaken up and this belief in itself,
or, himself, or, herself, is not destroyed (Step \ref{enu:A-belief-in-Self}).
This can suggest a hypothesis that what happens during our sleep might
be that our creators (Mother Nature or Evolution or Whatever), might
be giving suggestions to us in our sleep. These pointers for a better
life, (or greater intelligence, or, whatever their purpose, might
be), might manifest themselves as dreams. The Steps (\ref{enu:A-measure-to};
\ref{enu:A-method-to-reduce}) can be done in the background while
the agent is not necessarily asleep, but a period of complete focus
on the above two steps might be helpful.
\end{doublespace}
\end{enumerate}
\end{enumerate}
\begin{doublespace}
\textbf{\large{}End // The While Loop Ends Here, but it must go on
Forever}{\large\par}

The formal mathematical elements are discussed in Section (\ref{sec:From-Words-to-Symbols}).
These quantitative measures can be applied across aggregations of
smaller elements that can aid the AI agent by providing simple yet
powerful metrics to compare groups of entities and add to its knowledge
store. The results draw upon sources from statistics, probability,
marketing, economics / finance, communication systems, pattern recognition
and information theory. This becomes an example of how elements of
different fields can be combined to provide answers to the questions
raised by a particular field.
\end{doublespace}
\begin{doublespace}

\section{Conclusions and Possibilities for Future Research}
\end{doublespace}

\begin{doublespace}
We have discussed the intuition for why we need not just the best
computing science designers but also the best teachers to create artificial
intelligence. An unintended consequence of establishing curiosity
and confidence in an agent, expected to become intelligent, might
well be intelligence. We have considered why, even though we wish
to create intelligence and make the agent pass tests of intelligence,
the gift of intelligence might be something from the realm of the
unintentional. We have provided the mathematical tools and formal
elements of what such an endeavor might require, which includes models
of diffusion, distance measures and dimension reduction, among other
things.
\end{doublespace}
\begin{enumerate}
\begin{doublespace}
\item The possibilities for what improvements are necessary are endless
since we are just beginning. But once we instill confidence in the
agent that has to become intelligent, it can ask the questions to
learn better answers. That is, we try something, observe the mistakes
and make corrections depending on the level of progress we deem satisfactory. 
\item Another important aspect is to try to establish intelligence in simpler
real life applications. We then take the lessons to the more complex
design of a completely autonomous intelligent creature. There is a
lot of activity in this space on many individual fronts. We could
start with the financial markets, doing rudimentary household tasks,
driving etc. (all of which are happening). But combining the trial
and errors from all of these experiences are essential toward our
greater goal of AI. It would be helpful to start with the existing
level of understanding and the latest developments in text parsing,
speech recognition, and other areas. As we put these parts together,
the loop of trying and learning from mistakes has to continue forever
(or at-least for a very long time). 
\item As the likelihood of having to co-exist with so called artificially
created intelligent beings increases, we will need to learn to be
tolerant. We will need to focus on what truly makes us human and realize
the greater potential of our existence.
\end{doublespace}
\end{enumerate}
\begin{doublespace}
(Nilsson 2006) argues for the development of general-purpose educable
systems, that can learn and be taught to perform any of the thousands
of jobs that humans can perform, rather than work toward the goal
of automation by building special-purpose systems. But the message
we put forth is that the lessons from seemingly trivial tasks need
to be weaved towards higher ambitions. (Russell, Dewey \& Tegmark
2015) has some examples for further areas of research in building
intelligent agents. (Bottou 2014) suggests that, instead of trying
to bridge the gap between machine learning systems and sophisticated
“all-purpose” inference mechanisms, we can instead algebraically enrich
the set of manipulations applicable to training systems and build
reasoning capabilities from the ground up.

(Lake, etal 2017) review progress in cognitive science suggesting
that truly human-like learning and thinking machines will have to
reach beyond current engineering trends in both what they learn and
how they learn it. Specifically, they argue that these machines should
build causal models of the world, including intuitive theories of
physics and psychology, that support explanation and understanding
rather than merely solving pattern recognition problems. (Yannakakis
\& Togelius 2015) give a high-level overview of the field of artificial
and computational intelligence in games. (Chesani, Mello \& Milano
2017) propose to solve mathematical puzzles by means of computers,
starting from text and diagrams describing them, without any human
intervention.

The limited success in creating artificial intelligence, in machines,
humans and elsewhere as of today, is due to fundamental limitations
with the current thinking and the absence of certain basic principles
in the majority of attempts in this space right now. There are numerous
journals, articles and scientific efforts aimed at creating artificial
intelligence. The limited success can be attributed to many building
blocks of intelligence being absent in those efforts.
\end{doublespace}

This paper seeks to provide the foundational elements for intelligence
and addresses the drawbacks with the present efforts. Hence, the basic
ideas outlined in the paper will and should be of interest to anyone
interested in creating intelligence. While it is tempting to view
the topics presented here as being extremely diverse, we need to remind
ourselves that many instances of what we appreciate as intelligence
is a result of a demonstrated link between seemingly disparate elements
from wide ranging themes. Not to mention, as discussed in the introduction
(Section \ref{sec:The-Benchmark-for}), the disciplinary boundaries
we have created are artificial. For intelligence to happen such unnecessary
barriers have to be broken down.

\begin{doublespace}
Surely, one paper cannot completely accomplish the task where millions
of other efforts have failed. But what it can hope to do is guide
future efforts to areas that will lead to greater success, inspired
by how intelligence perhaps happens in us. Having said that, a road-map
which can act as an algorithm and also a set of hypothesis that can
be tested and implemented computationally are provided. Just because
we (readers / reviewers / students) do not see a connection or do
not understand something, does not mean there is no connection or
nothing to be understood. This paper puts forth the suggestion that,
in such cases when something is not clear, to increase intelligence
we need to ask questions and relate it to what we already know. Otherwise,
intelligence will not increase. This message is constantly espoused
in classrooms worldwide. But when we deem ourselves above that, whatever
our role in the intelligence creation eco-system, we have stopped
to learn. This paper is as much about intelligence in humans as in
machines. Intelligence happens as an unintended consequence due to
curiosity left free. If something seems irrelevant, then we have stopped
being curious and the level of intelligence plateaus off.

We have a great example of intelligent beings that have been created,
which is us. We could debate as to whether this creator is evolution
or an intelligent designer. But until higher powers intervene and
provide the ultimate solution to create intelligence. We have to make
do with marginal methods, exemplified by curiously confident, trials
and errors, such as this composition puts forth.

As we wait for the perfect solution, it is worth meditating upon what
superior beings would do when faced with a complex situation, such
as the one we are in. It is said that the Universe is but the Brahma's
(Creator's) dream (Barnett 1907; Ramamurthi 1995; Ghatage 2010). Research
(Effort / Struggle) can help us understand this world and maybe decipher
the key to intelligent agents. Sleep (Ease / Peace of Mind) can help
us create our own world. Surely, creating intelligent beings would
be a much smaller part of this new world. Also, with very little doubt,
sleep has many direct benefits to increase cognitive abilities. We
just need to be mindful that the most rosy and well intentioned dreams
can have unintended consequences and turn to nightmares (Nolan 2010;
Lehrer 2010; Kashyap 2016).

Native to Australia (Clark 1993), ``Koalas spend about 4.7 hours
eating, 4 minutes traveling, 4.8 hours resting while awake and 14.5
hours sleeping in a 24-hour period'' (Nagy \& Martin 1985; Smith
1979; Moyal 2008). The benefits of yoga on sleep quality are well
documented (Cohen, etal 2004; Khalsa 2004; Manjunath \& Telles 2005;
Chen, etal 2009; Vera, etal 2009; End-note \ref{enu:Yoga}).

A lesson from close by and down under: ``We need to Do Some Yoga
and Sleep Like A Koala”.
\end{doublespace}
\begin{doublespace}

\section{End-notes}
\end{doublespace}
\begin{enumerate}
\begin{doublespace}
\item \label{enu:Changing-D=000026A-which}Changing D\&A, which gives rise
to different Q\&A, might even be telling us that Q\&A and D\&A might
be in our very DNA, the biological one, which are always changing
(Alberts, etal 2002; End-notes \ref{DNA}; \ref{enu:Maybe,-DNA-hold}).
\item \label{DNA}Deoxyribonucleic acid (DNA) is a molecule composed of
two chains (made of nucleotides) that coil around each other to form
a double helix carrying the genetic instructions used in the growth,
development, functioning, and reproduction of all known living organisms
and many viruses. DNA and ribonucleic acid (RNA) are nucleic acids.
Alongside proteins, lipids and complex carbohydrates (polysaccharides),
nucleic acids are one of the four major types of macromolecules that
are essential for all known forms of life. \href{https://en.wikipedia.org/wiki/DNA}{DNA, Wikipedia Link}
\item \label{enu:Maybe,-DNA-hold}Maybe, DNA hold the lessons from the lives
of every ancestor we have ever had. Evolution is constantly coding
the information, compressing it and passing forward what is needed
to survive better and to thrive, building what is essential right
into our genes. For information storage in DNA and related applications
see: Church, Gao \& Kosuri 2012; Lutz, etal 2013; Kosuri \& Church
2014; Roy, etal 2015. 
\item \label{enu:Art-Science}A frame of mind and approach to seeking knowledge
that is open to the methods of both science and art could be termed,
``Science without Borders but Combined with the Arts''.
\item \label{enu:Universal Computing Machine}In computer science, a universal
Turing machine (UTM) is a Turing machine (Minsky 1967; End-note \ref{Turing-Machine})
that can simulate an arbitrary Turing machine on arbitrary input.
The universal machine essentially achieves this by reading both the
description of the machine to be simulated as well as the input thereof
from its own tape. \href{https://en.wikipedia.org/wiki/Universal_Turing_machine}{Universal Computing Machine, Wikipedia Link}
\item \label{Turing-Machine}A Turing machine is a mathematical model of
computation that defines an abstract machine, which manipulates symbols
on a strip of tape according to a table of rules. Despite the model's
simplicity, given any computer algorithm, a Turing machine capable
of simulating that algorithm's logic can be constructed (Sipser 2006).
\href{https://en.wikipedia.org/wiki/Turing_machine}{Turing Machine, Wikipedia Link}
\item \label{enu:Computer}A computer is a device that can be instructed
to carry out sequences of arithmetic or logical operations automatically
via computer programming. \href{https://en.wikipedia.org/wiki/Computer}{Computer, Wikipedia Link}
\item \label{enu:MAC vs MPC}Apple Computer, Inc. v. Microsoft Corporation,
was a copyright infringement lawsuit in 1994 in which Apple Computer,
Inc. (now Apple Inc.) sought to prevent Microsoft and Hewlett-Packard
from using visual graphical user interface (GUI) elements that were
similar to those in Apple's Lisa and Macintosh operating systems.
Mac vs PC also refers to the rivalry between the two companies to
dominate the personal computer market. \href{https://en.wikipedia.org/wiki/Apple_Computer,_Inc._v._Microsoft_Corp.}{MAC vs MPC, Wikipedia Link}
\item \label{enu:History Computing}The history of computing is longer than
the history of computing hardware and modern computing technology
and includes the history of methods intended for pen and paper or
for chalk and slate, with or without the aid of tables. \href{https://en.wikipedia.org/wiki/History_of_computing}{History of Computing, Wikipedia Link}
\item \label{enu:Mother-Daughter}To illustrate the grammatical ambiguities
that exist (persist?) in many modern languages, consider this example:
``A mother beats up her daughter because she was drunk''. So, who
was the drunk person in this incident? \href{https://ell.stackexchange.com/questions/155336/how-to-interpret-the-sentence-a-mother-beats-up-her-daughter-because-she-was}{Mother Beats Daughter, English Language Learners Link}
\item \label{enu:Beauty-Beast}Beauty and the Beast (French: La Belle et
la Bête) is a fairy tale written by French novelist Gabrielle-Suzanne
Barbot de Villeneuve and published in 1740 in The Young American and
Marine Tales (French: La Jeune Américaine et les contes marins). Her
lengthy version was abridged, rewritten, and published first by Jeanne-Marie
Leprince de Beaumont in 1756. \href{https://en.wikipedia.org/wiki/Beauty_and_the_Beast}{Beauty and the Beast, Wikipedia Link}
\item \label{enu:The-Turing-test}The Turing test, developed by Alan Turing
in 1950, is a test of a machine's ability to exhibit intelligent behavior
equivalent to, or indistinguishable from, that of a human. Turing
proposed that a human evaluator would judge natural language conversations
between a human and a machine designed to generate human-like responses.
The evaluator would be aware that one of the two partners in conversation
is a machine, and all participants would be separated from one another.
The conversation would be limited to a text-only channel such as a
computer keyboard and screen so the result would not depend on the
machine's ability to render words as speech (Turing originally suggested
a teleprinter, one of the few text-only communication systems available
in 1950). If the evaluator cannot reliably tell the machine from the
human, the machine is said to have passed the test. The test does
not check the ability to give correct answers to questions, only how
closely answers resemble those a human would give. \href{https://en.wikipedia.org/wiki/Turing_test}{Turing Test, Wikipedia Link}
\item \label{enu:Chinese room thought experiment}Searle's thought experiment
begins with this hypothetical premise: suppose that artificial intelligence
research has succeeded in constructing a computer that behaves as
if it understands Chinese. It takes Chinese characters as input and,
by following the instructions of a computer program, produces other
Chinese characters, which it presents as output. Suppose, says Searle,
that this computer performs its task so convincingly that it comfortably
passes the Turing test: it convinces a human Chinese speaker that
the program is itself a live Chinese speaker. To all of the questions
that the person asks, it makes appropriate responses, such that any
Chinese speaker would be convinced that they are talking to another
Chinese-speaking human being.
\end{doublespace}

\begin{doublespace}
This was originally phrased as: Searle supposes that he is in a closed
room and has a book with an English version of the computer program,
along with sufficient paper, pencils, erasers, and filing cabinets.
Searle could receive Chinese characters through a slot in the door,
process them according to the program's instructions, and produce
Chinese characters as output. 

The question Searle wants to answer is this: does the machine literally
\textquotedbl understand\textquotedbl{} Chinese? Or is it merely
simulating the ability to understand Chinese? Searle calls the first
position \textquotedbl strong AI\textquotedbl{} and the latter \textquotedbl weak
AI\textquotedbl . Searle writes that \textquotedbl according to
Strong AI, the correct simulation really is a mind. According to Weak
AI, the correct simulation is a model of the mind.\textquotedbl{}
He also writes: \textquotedbl On the Strong AI view, the appropriately
programmed computer does not just simulate having a mind; it literally
has a mind.\textquotedbl{} \href{https://en.wikipedia.org/wiki/Chinese_room\#Chinese_room_thought_experiment}{Searle's Chinese room thought experiment, Wikipedia Link} 
\end{doublespace}
\begin{doublespace}
\item \label{enu:Certain-six-year}Certain six year old's that we know when
questioned, ``How to make computers intelligent?'', responded by
saying, ``Have two computers. Use Google on one computer to find
the answer and make the other computer use this answer''. This remark
gives us assurance that the concepts put forth in this paper about
the way children learn by being curious, confident and most importantly
by imitating good role models while attempting to solve problems,
without getting frazzled, hold the key to increased intelligence.
\item \label{enu:Taleb and Kahneman discuss Trial and Error / IQ Points}As
Taleb explains, ``it is trial with small errors that leads to progress''.
That said, there could be big errors that might incapacitate the person
trying the trial from attempting further trials. But as long as someone
else has observed the attempts with huge errors, the rest of society
benefits from it. We need to assume, of course, that the big blow
up has left a substantial portion of society intact, or at-least not
too shaken up. This concept is also illustrated in Point (\ref{enu:Lessons-from-other};
Section \ref{subsec:Becoming-Smarter-than}) about learning from the
lessons history holds for us. (Ismail 2014) mentions the following
quote from Taleb, “Knowledge gives you a little bit of an edge, but
tinkering (trial and error) is the equivalent of 1000 IQ points. It
is tinkering that allowed the industrial revolution''. This means
that to match trial and error we need 1000 IQ points. But trial and
error could still give the wrong outcomes. We can try and fail many
times, never finding the right answers, and still be wrong. So in
our paper we make the assumption that we need 2000 IQ points to consistently
make the right decisions. The subtle point that arises from this discussion
is that we need 2000 IQ points to be right all the time, but the problem
is that the best of us has less than 200 IQ points. \href{https://www.youtube.com/watch?v=MMBclvY_EMA}{Nassim Taleb and Daniel Kahneman discuss Trial and Error / IQ Points, among other things, at the New York Public Library on Feb 5, 2013. (Link)} 
\item \label{enu:Bass-Model-Diffusion}The Bass Model or Bass Diffusion
Model was developed by Frank Bass. It consists of a simple differential
equation that describes the process of how new products get adopted
in a population. The model presents a rationale of how current adopters
and potential adopters of a new product interact. The basic premise
of the model is that adopters can be classified as innovators or as
imitators and the speed and timing of adoption depends on their degree
of innovativeness and the degree of imitation among adopters. The
Bass model has been widely used in forecasting, especially new products'
sales forecasting and technology forecasting. Mathematically, the
basic Bass diffusion is a Riccati equation (End-note \ref{enu:Ricatti-Equation})
with constant coefficients. \href{https://en.wikipedia.org/wiki/Bass_diffusion_model}{Bass Model of Diffusion, Wikipedia Link} 
\item \label{enu:Ricatti-Equation}In mathematics, a Riccati equation in
the narrowest sense is any first-order ordinary differential equation
that is quadratic in the unknown function. In other words, it is an
equation of the form 
\[
y'(x)=q_{0}(x)+q_{1}(x)\,y(x)+q_{2}(x)\,y^{2}(x)
\]
 where $q_{0}(x)\neq0$ and $q_{2}(x)\neq0$. If $q_{0}(x)=0$ the
equation reduces to a Bernoulli equation, while if $q_{2}(x)=0$ the
equation becomes a first order linear ordinary differential equation.
The equation is named after Jacopo Riccati (1676–1754) (see Riccati
1724). \href{https://en.wikipedia.org/wiki/Riccati_equation}{Riccati Equation, Wikipedia Link} 
\end{doublespace}
\item \label{enu:Infinite-Progress}(Kashyap 2019b) provides an infinite
progress benchmark to be successful and has a detailed discussion.
This measure of success is related to taking the first step.
\begin{doublespace}
\item \label{enu:Yoga} Yoga is a group of physical, mental, and spiritual
practices which originated in ancient India. \href{https://en.wikipedia.org/wiki/Yoga}{Yoga, Wikipedia Link}
\item \label{enu:The-more-curious-baby}The more curious a child is, the
more he learns. Nurturing your child’s curiosity is one of the most
important ways you can help her become a lifelong learner. Babies
are born learners, with a natural curiosity to figure out how the
world works. Curiosity is the desire to learn. It is an eagerness
to explore, discover and figure things out. \href{https://www.zerotothree.org/resources/224-tips-on-nurturing-your-child-s-curiosity}{Tips on Nurturing Your Child's Curiosity,  Link} 
\end{doublespace}
\end{enumerate}
\begin{doublespace}

\section{\label{sec:References}References}
\end{doublespace}
\begin{enumerate}
\begin{doublespace}
\item Alberts, B., Johnson, A., Lewis, J., Raff, M., Roberts, K., \& Walter,
P. (2002). Molecular Biology of the Cell, Garland Science, New York.
\item Aly, M., \& Moscovitch, M. (2010). The effects of sleep on episodic
memory in older and younger adults. Memory, 18(3), 327-334.
\item Alpaydin, E. (2014). Introduction to machine learning. MIT press.
\item Armbrust, M., Fox, A., Griffith, R., Joseph, A. D., Katz, R., Konwinski,
A., ... \& Zaharia, M. (2010). A view of cloud computing. Communications
of the ACM, 53(4), 50-58.
\item Amir, Y., Ben-Ishay, E., Levner, D., Ittah, S., Abu-Horowitz, A.,
\& Bachelet, I. (2014). Universal computing by DNA origami robots
in a living animal. Nature nanotechnology, 9(5), 353.
\item Banerjee, A. V. (1992). A simple model of herd behavior. The Quarterly
Journal of Economics, 107(3), 797-817.
\item Banerjee, A. V. (1993). The economics of rumours. The Review of Economic
Studies, 60(2), 309-327.
\item Barnett, L. D. (1907). The Brahma Knowledge. An Outline of the Philosophy
of the Vedanta as Set Forth by the Upanishads and by Sankara. Wisdom
of the East series. E.P. Dutton Publishing, Boston, Massachusetts.
\item Bartholomew, D. J. (2004). Measuring intelligence: Facts and fallacies.
Cambridge University Press.
\item Bass, F. M. (1969). A new product growth for model consumer durables.
Management science, 15(5), 215-227.
\item Bass, F. M., Krishnan, T. V., \& Jain, D. C. (1994). Why the Bass
model fits without decision variables. Marketing science, 13(3), 203-223.
\item Beck, J. V., \& Arnold, K. J. (1977). Parameter estimation in engineering
and science. James Beck.
\item Bénabou, R., \& Tirole, J. (2002). Self-confidence and personal motivation.
The Quarterly Journal of Economics, 117(3), 871-915.
\item Berlyne, D. E. (1954). A theory of human curiosity. British Journal
of Psychology, 45, 180-191.
\item Berlyne, D. E. (1966). Curiosity and exploration. Science, 153. 25-33.
\item Bhattacharyya, A. (1943). On a Measure of Divergence Between Two Statistical
Populations Defined by their Probability Distributions, Bull. Calcutta
Math. Soc., 35, pp. 99-110.
\item Bhattacharyya, A. (1946). On a measure of divergence between two multinomial
populations. Sankhyā: The Indian Journal of Statistics, 401-406.
\item Blischke, K., Erlacher, D., Kresin, H., Brueckner, S., \& Malangré,
A. (2008). Benefits of sleep in motor learning–prospects and limitations.
Journal of human kinetics, 20, 23-35.
\item Bottou, L. (2014). From machine learning to machine reasoning. Machine
learning, 94(2), 133-149.
\item Brooks, R. A. (1991). Intelligence without representation. Artificial
intelligence, 47(1-3), 139-159.
\item Burges, C. J. (2009). Dimension reduction: A guided tour. Machine
Learning, 2(4), 275-365.
\item Burkardt, J. (2014). The Truncated Normal Distribution. Department
of Scientific Computing Website, Florida State University.
\item Bush, R. R., \& Mosteller, F. (2006). A mathematical model for simple
learning. In Selected Papers of Frederick Mosteller (pp. 221-234).
Springer New York.
\item Bush, R. R., \& Mosteller, F. (1955). Stochastic models for learning.
\item Calkins, L., \& Bellino, L. (1997). Raising Lifelong Learners: A Parent's
Guide. Harper Collins, 1000 Keystone Industrial Park, Scranton, PA
18512; toll-free.
\item Cameron, J., \& Wisher, W. (1991). Terminator 2: judgment day (Vol.
2). USA.
\item Campbell, D. T. (1956). Perception as substitute trial and error.
Psychological review, 63(5), 330.
\item Campbell-Kelly, M. (2001). Not only Microsoft: The maturing of the
personal computer software industry, 1982–1995. Business History Review,
75(1), 103-145.
\item Carlton, J., \& Annotations-Kawasaki, G. (1997). Apple: The inside
story of intrigue, egomania, and business blunders. Random House Inc..
\item Castelvecchi, D. (2016). Can we open the black box of AI?. Nature,
538(7623), 20-23.
\item Catania, B., \& Zarri, G. P. (2000). Intelligent database systems.
Addison-Wesley.
\item Ceci, S. J., \& Liker, J. K. (1986). A day at the races: A study of
IQ, expertise, and cognitive complexity. Journal of Experimental Psychology:
General, 115(3), 255.
\item Ceruzzi, P. E. (2003). A history of modern computing. MIT press.
\item Chen, K. M., Chen, M. H., Chao, H. C., Hung, H. M., Lin, H. S., \&
Li, C. H. (2009). Sleep quality, depression state, and health status
of older adults after silver yoga exercises: cluster randomized trial.
International journal of nursing studies, 46(2), 154-163.
\item Chesani, F., Mello, P., \& Milano, M. (2017). Solving Mathematical
Puzzles: A Challenging Competition for AI. AI Magazine, 38(3), 83-97.
\item Chiani, M., Dardari, D., \& Simon, M. K. (2003). New exponential bounds
and approximations for the computation of error probability in fading
channels. Wireless Communications, IEEE Transactions on, 2(4), 840-845.
\item Church, G. M., Gao, Y., \& Kosuri, S. (2012). Next-generation digital
information storage in DNA. Science, 1226355.
\item Clark, M. (1993). History of Australia. Melbourne University Publish.
\item Clément, R., Dörnyei, Z., \& Noels, K. A. (1994). Motivation, self‐confidence,
and group cohesion in the foreign language classroom. Language learning,
44(3), 417-448.
\item Cody, W. J. (1969). Rational Chebyshev approximations for the error
function. Mathematics of Computation, 23(107), 631-637.
\item Cohen, L., Warneke, C., Fouladi, R. T., Rodriguez, M., \& Chaoul‐Reich,
A. (2004). Psychological adjustment and sleep quality in a randomized
trial of the effects of a Tibetan yoga intervention in patients with
lymphoma. Cancer, 100(10), 2253-2260.
\item Coogan, P. (2009). The Definition of the Superhero. A comics studies
reader, 77.
\item Copeland, B. J. (2000). The turing test. Minds and Machines, 10(4),
519-539.
\item Corcoran, P., Coughlin, T., \& Wozniak, S. (2016). Champions in our
midst: the Apple doesn't fall far from the tree. IEEE Consumer Electronics
Magazine, 5(1), 93-98.
\item Darwin, C. (1859). On the origin of species by means of natural selection.
1968. London: Murray Google Scholar.
\item Dasgupta, S., \& Gupta, A. (1999). An elementary proof of the Johnson-Lindenstrauss
lemma. International Computer Science Institute, Technical Report,
99-006.
\item Davis, M. (2011). The universal computer: The road from Leibniz to
Turing. AK Peters/CRC Press.
\item Dawkins, R. (1976). The selfish gene. Oxford university press.
\item De Beaumont, M. L. P. (1804). Beauty and the Beast. Prabhat Prakashan.
\item De Brigard, F. (2014). The nature of memory traces. Philosophy Compass,
9(6), 402-414.
\item DeDonno, M. A. (2016). The influence of IQ on pure discovery and guided
discovery learning of a complex real-world task. Learning and Individual
Differences, 49, 11-16.
\item Deng, L., \& Yu, D. (2014). Deep learning: methods and applications.
Foundations and Trends® in Signal Processing, 7(3–4), 197-387.
\item Denning, P. J. (2005). Is computer science science?. Communications
of the ACM, 48(4), 27-31.
\item Derpanis, K. G. (2008). The Bhattacharyya Measure. Mendeley Computer,
1(4), 1990-1992.
\item Doidge, N. (2007). The brain that changes itself: stories of personal
triumph from the frontiers of brain science/Norman.
\item Durant, W. (1968). The lessons of history.
\item Eco, U., \& Chilton, N. (1972). The myth of Superman.
\item Eldredge, N. (2005). Darwin: discovering the tree of life. WW Norton
\& Company.
\item Ellenbogen, J. M. (2005). Cognitive benefits of sleep and their loss
due to sleep deprivation. Neurology, 64(7), E25-E27.
\item Feltz, D. L. (1988). Self-confidence and sports performance. Exercise
and sport sciences reviews, 16(1), 423-458.
\item Fingeroth, D. (2004). Superman on the Couch: What Superheroes Really
Tell Us about Ourselves and Our Society. A\&C Black.
\item Fodor, I. K. (2002). A survey of dimension reduction techniques. Technical
Report UCRL-ID-148494, Lawrence Livermore National Laboratory.
\item Fogel, A. (2004). Infancy: Accessing Our Earliest Experiences. Theories
of infant development, 204.
\item Frankl, P., \& Maehara, H. (1988). The Johnson-Lindenstrauss lemma
and the sphericity of some graphs. Journal of Combinatorial Theory,
Series B, 44(3), 355-362.
\item Frankl, P., \& Maehara, H. (1990). Some geometric applications of
the beta distribution. Annals of the Institute of Statistical Mathematics,
42(3), 463-474.
\item Freiberger, P., \& Swaine, M. (1999). Fire in the Valley: the making
of the personal computer. McGraw-Hill Professional.
\item French, R. M. (1990). Subcognition and the limits of the Turing test.
Mind, 99(393), 53-65.
\item French, R. M. (2000). The Turing Test: the first 50 years. Trends
in cognitive sciences, 4(3), 115-122.
\item Gamble, W. C., \& Cota-Robles, S. (2015). Guiding Curiosity: Nurturing
Young Scientists. BookBaby.
\item Garland, H. (1977). Design innovations in personal computers. Computer,
10(3), 24-27.
\item Ghahramani, Z. (2015). Probabilistic machine learning and artificial
intelligence. Nature, 521(7553), 452-460.
\item Ghatage, S. (2010). Brahma's Dream. Anchor Canada, Penguin Random
House, Manhattan, New York.
\item Gibbons, R. (1992). A primer in game theory. Harvester Wheatsheaf.
\item Giles, J. (2005). Wisdom of the Crowd. Nature, 438(7066), 281.
\item Gopnik, A., Meltzoff, A. N., \& Kuhl, P. K. (1999). The scientist
in the crib: Minds, brains, and how children learn. William Morrow
\& Co.
\item Guo, T. (2015). Alan Turing: Artificial intelligence as human self‐knowledge.
Anthropology Today, 31(6), 3-7.
\item Hagen, U. (1991). Challenge for the Actor. Simon and Schuster.
\item Hagen, U. (1973). Respect for acting. John Wiley \& Sons.
\item Halpern, J. H. (2004). Hallucinogens and dissociative agents naturally
growing in the United States. Pharmacology \& therapeutics, 102(2),
131-138.
\item Harnad, S. (1992). The Turing Test is not a trick: Turing indistinguishability
is a scientific criterion. ACM SIGART Bulletin, 3(4), 9-10.
\item Haslem, W., Ndalianis, A., \& Mackie, C. J. (Eds.). (2007). Super/Heroes:
From Hercules to Superman. New Academia Publishing, LLC.
\item Haykin, S. S. (2004). Neural networks: A comprehensive foundation.
\item Haykin, S. S. (2009). Neural networks and learning machines (Vol.
3). Upper Saddle River, NJ, USA:: Pearson.
\item Hernández-Orallo, J., \& Dowe, D. L. (2010). Measuring universal intelligence:
Towards an anytime intelligence test. Artificial Intelligence, 174(18),
1508-1539.
\item Hernández-Orallo, J., Martínez-Plumed, F., Schmid, U., Siebers, M.,
\& Dowe, D. L. (2016). Computer models solving intelligence test problems:
Progress and implications. Artificial Intelligence, 230, 74-107.
\item Holt, J. (2017). How Children Learn. Classics in Child Development.
\item Horrace, W. C. (2005). Some results on the multivariate truncated
normal distribution. Journal of Multivariate Analysis, 94(1), 209-221.
\item Howe, M. L., \& Courage, M. L. (1993). On resolving the enigma of
infantile amnesia. Psychological bulletin, 113(2), 305.
\item Ifrah, G., Harding, E. F., Bellos, D., \& Wood, S. (2000). The universal
history of computing: From the abacus to quantum computing. John Wiley
\& Sons, Inc.
\item Ismail, S. (2014). Exponential Organizations: Why new organizations
are ten times better, faster, and cheaper than yours (and what to
do about it). Diversion Books.
\item Jacobs, P. S. (2014). Text-based intelligent systems: Current research
and practice in information extraction and retrieval. Psychology Press.
\item Jiang, Z., \& Jain, D. C. (2012). A generalized Norton–Bass model
for multigeneration diffusion. Management Science, 58(10), 1887-1897.
\item Johnson, W. B., \& Lindenstrauss, J. (1984). Extensions of Lipschitz
mappings into a Hilbert space. Contemporary mathematics, 26(189-206),
1.
\item Johnson, J. S., \& Newport, E. L. (1989). Critical period effects
in second language learning: The influence of maturational state on
the acquisition of English as a second language. Cognitive psychology,
21(1), 60-99.
\item Jones, N. (2014). The learning machines. Nature, 505(7482), 146.
\item Kashyap, R. (2016). Notes on Uncertainty, Unintended Consequences
and Everything Else. Working Paper.
\item Kashyap, R. (2017). Imitation in the Imitation Game. Working Paper. 
\end{doublespace}
\item Kashyap, R. (2018). Seven Survival Senses: Evolutionary Training makes
Discerning Differences more Natural than Spotting Similarities. World
Futures, Accepted, Forthcoming.
\item Kashyap, R. (2018b). The Brain is in the Head, But The Mind is in
the Belly: Food For Thought?. Working Paper.
\begin{doublespace}
\item Kashyap, R. (2019). The Perfect Marriage and Much More: Combining
Dimension Reduction, Distance Measures and Covariance. Physica A:
Statistical Mechanics and its Applications, 536, 120938.
\end{doublespace}
\item Kashyap, R. (2019b). For Whom the Bell (Curve) Tolls: A to F, Trade
Your Grade Based on the Net Present Value of Friendships with Financial
Incentives. The Journal of Private Equity, 22(3), 64-81. Chicago 
\begin{doublespace}
\item Kattumannil, S. K. (2009). On Stein’s identity and its applications.
Statistics \& Probability Letters, 79(12), 1444-1449.
\item Keynes, J. M. (1937). The General Theory of Employment. The Quarterly
Journal of Economics, 51(2), 209-223.
\item Keynes, J. M. (1971). The Collected Writings of John Maynard Keynes:
In 2 Volumes. A Treatise on Money. The Applied Theory of Money. Macmillan
for the Royal Economic Society. 
\item Keynes, J. M. (1973). A treatise on probability, the collected writings
of John Maynard Keynes, vol. VIII.
\item Khalsa, S. B. S. (2004). Treatment of chronic insomnia with yoga:
A preliminary study with sleep–wake diaries. Applied psychophysiology
and biofeedback, 29(4), 269-278.
\item Kiani, M., Panaretos, J., Psarakis, S., \& Saleem, M. (2008). Approximations
to the normal distribution function and an extended table for the
mean range of the normal variables.
\item Kimeldorf, G., \& Sampson, A. (1973). A class of covariance inequalities.
Journal of the American Statistical Association, 68(341), 228-230.
\item Kosuri, S., \& Church, G. M. (2014). Large-scale de novo DNA synthesis:
technologies and applications. Nature methods, 11(5), 499. 
\item Kuhl, P. K. (2004). Early language acquisition: cracking the speech
code. Nature Reviews. Neuroscience, 5(11), 831.
\item Lake, B. M., Ullman, T. D., Tenenbaum, J. B., \& Gershman, S. J. (2017).
Building machines that learn and think like people. Behavioral and
Brain Sciences, 40.
\item Lawson, T. (1985). Uncertainty and economic analysis. The Economic
Journal, 95(380), 909-927.
\item LeBlanc, M. D., \& Weber‐Russell, S. (1996). Text integration and
mathematical connections: A computer model of arithmetic word problem
solving. Cognitive Science, 20(3), 357-407.
\item LeCun, Y., Bengio, Y., \& Hinton, G. (2015). Deep learning. nature,
521(7553), 436-444.
\item Legg, S., \& Hutter, M. (2007). Universal intelligence: A definition
of machine intelligence. Minds and Machines, 17(4), 391-444.
\item Lee, K. Y., \& Bretschneider, T. R. (2012). Separability Measures
of Target Classes for Polarimetric Synthetic Aperture Radar Imagery.
Asian Journal of Geoinformatics, 12(2).
\item Lehrer, J. (2010). \href{https://www.wired.com/2010/07/the-neuroscience-of-inception/}{The Neuroscience of Inception}.
Wired 26 Jul. 2010. Web. 13 Aug. 2013.
\item Lenneberg, E. H. (1967). The biological foundations of language. Hospital
Practice, 2(12), 59-67.
\item Leuenberger, M. N., \& Loss, D. (2001). Quantum computing in molecular
magnets. Nature, 410(6830), 789.
\item Lightbown, P. M., Spada, N., Ranta, L., \& Rand, J. (1999). How languages
are learned (Vol. 2). Oxford: Oxford University Press.
\item Litman, J. A., \& Spielberger, C. D. (2003). Measuring epistemic curiosity
and its diversive and specific components. Journal of Personality
Assessment, 80, 75-86.
\item Loewenstein, G. (1994). The psychology of curiosity: A review and
reinterpretation. Psychological bulletin, 116(1), 75.
\item Loewy, E. H. (1998). Curiosity, imagination, compassion, science and
ethics: Do curiosity and imagination serve a central function?. Health
Care Analysis, 6(4), 286-294.
\item Lutz, J. F., Ouchi, M., Liu, D. R., \& Sawamoto, M. (2013). Sequence-controlled
polymers. Science, 341(6146), 1238149.
\item Lynch, T. J. (1985). Data compression: techniques and applications.
Lifetime Learning Publications.
\item Mahajan, V. (1985). Innovation diffusion. John Wiley \& Sons, Ltd.
\item Mahajan, V., Muller, E., \& Bass, F. M. (1991). New product diffusion
models in marketing: A review and directions for research. In Diffusion
of technologies and social behavior (pp. 125-177). Springer, Berlin,
Heidelberg.
\item Malomo, A. O., Idowu, O. E., \& Osuagwu, F. C. (2006). Lessons from
history: human anatomy, from the origin to the renaissance. Int. J.
Morphol, 24(1), 99-104.
\item Manes, S., \& Andrews, P. (1993). Gates: How Microsoft's mogul reinvented
an industry-and made himself the richest man in America. Simon \&
Schuster.
\item Manjunath, N. K., \& Telles, S. (2005). Influence of Yoga \& Ayurveda
on self-rated sleep in a geriatric population. Indian Journal of Medical
Research, 121(5), 683.
\item Martínez-Plumed, F., Ferri, C., Hernández-Orallo, J., \& Ramírez-Quintana,
M. J. (2017). A computational analysis of general intelligence tests
for evaluating cognitive development. Cognitive Systems Research,
43, 100-118.
\item Mazur, J. E. (2015). Learning and behavior. Psychology Press.
\item McCarthy, J., \& Hayes, P. J. (1969). Some philosophical problems
from the standpoint of artificial intelligence. Readings in artificial
intelligence, 431-450.
\item McCarthy, J. (2004). What is artificial intelligence. URL: http://www-formal.
stanford. edu/jmc/whatisai. html.
\item McCulloch, W. S., \& Pitts, W. (1943). A logical calculus of the ideas
immanent in nervous activity. The bulletin of mathematical biophysics,
5(4), 115-133.
\item McManus, H., \& Hastings, D. (2005, July). 3.4. 1 A Framework for
Understanding Uncertainty and its Mitigation and Exploitation in Complex
Systems. In INCOSE International Symposium (Vol. 15, No. 1, pp. 484-503).
\item Michalakelis, C., Varoutas, D., \& Sphicopoulos, T. (2010). Innovation
diffusion with generation substitution effects. Technological Forecasting
and Social Change, 77(4), 541-557.
\item Mill, J. (1829). Analysis of the Phenomena of the Human Mind (Vol.
1, 2). Longmans, Green, Reader, and Dyer.
\item Minsky, M. L. (1967). Computation: finite and infinite machines. Prentice-Hall,
Inc.
\item Moor, J. H. (1976). An analysis of the Turing test. Philosophical
Studies, 30(4), 249-257.
\item Moyal, A. (Ed.). (2008). Koala: a historical biography. CSIRO PUBLISHING.
\item Musshoff, F., Madea, B., \& Beike, J. (2000). Hallucinogenic mushrooms
on the German market—simple instructions for examination and identification.
Forensic science international, 113(1), 389-395.
\item Nadel, L., \& Zola-Morgan, S. (1984). Infantile amnesia. In Infant
memory (pp. 145-172). Springer US.
\item Nagy, K. A., \& Martin, R. W. (1985). Field Metabolic Rate, Water
Flux, Food Consumption and Time Budget of Koalas, Phascolarctos Cinereus
(Marsupialia: Phascolarctidae) in Victoria. Australian Journal of
Zoology, 33(5), 655-665.
\item Nash, J. F. (1950). Equilibrium points in n-person games. Proceedings
of the national academy of sciences, 36(1), 48-49.
\item Nass, M. M., \& Cooper, L. N. (1975). A theory for the development
of feature detecting cells in visual cortex. Biological cybernetics,
19(1), 1-18.
\item Nere, A., Hashmi, A., Cirelli, C., \& Tononi, G. (2013). Sleep-dependent
synaptic down-selection (I): modeling the benefits of sleep on memory
consolidation and integration. Frontiers in neurology.
\item Newport, E. L. (1990). Maturational constraints on language learning.
Cognitive science, 14(1), 11-28.
\item Nilsson, N. J. (2006). Human-Level Artificial Intelligence? Be Serious!.
AI Magazine, 26(4), 68-75.
\item Niu, S. C. (2002). A stochastic formulation of the Bass model of new-product
diffusion. Mathematical problems in Engineering, 8(3), 249-263.
\item Noels, K. A., Pon, G., \& Clément, R. (1996). Language, identity,
and adjustment: The role of linguistic self-confidence in the acculturation
process. Journal of language and social psychology, 15(3), 246-264.
\item Nolan, C. (2010). Inception {[}film{]}. Warner Bros.: Los Angeles,
CA, USA.
\item Norstad, J. (1999). The normal and lognormal distributions.
\item Oja, E. (1982). Simplified neuron model as a principal component analyzer.
Journal of mathematical biology, 15(3), 267-273.
\item Okuda, S. M., Runco, M. A., \& Berger, D. E. (1991). Creativity and
the finding and solving of real-world problems. Journal of Psychoeducational
assessment, 9(1), 45-53.
\item Owens, T. J. (1993). Accentuate the positive-and the negative: Rethinking
the use of self-esteem, self-deprecation, and self-confidence. Social
Psychology Quarterly, 288-299.
\item Parsaye, K., \& Chignell, M. (1993). Intelligent Database Tools and
Applications: Hyperinformation access, data quality, visualization,
automatic discovery. John Wiley \& Sons, Inc..
\item Penrose, R. (1989). The Emperor’s New Mind: concerning computers,
brains and the laws of physics. Oxford Paperbacks.
\item Perrier, J. Y., Sipper, M., \& Zahnd, J. (1996). Toward a viable,
self-reproducing universal computer. Physica D: Nonlinear Phenomena,
97(4), 335-352.
\end{doublespace}
\item Phillips, F., Lin, H., Schifter, T., \& Folse, N. (2019). Augmented
Popperian Experiments: A Framework for Sustainability Knowledge Development
across Contexts. European J. of International Management, In Press,
DOI: 10.1504/EJIM.2020.10024697.
\begin{doublespace}
\item Piskorski, J., \& Neumann, G. (2000). An intelligent text extraction
and navigation system. In Content-Based Multimedia Information Access-Volume
2 (pp. 1015-1032).
\item Popper, K. R. (2002). The poverty of historicism. Psychology Press.
\item Porter, L. (2014). Benedict Cumberbatch, Transition Completed: Films,
Fame, Fans. Andrews UK Limited.
\item Preston, J., \& Bishop, M. J. (2002). Views into the Chinese room:
New essays on Searle and artificial intelligence. OUP.
\item Proudfoot, D. (2015). What turing himself said about the imitation
game. IEEE Spectrum, 52(7), 42-47.
\item Ramamurthi, B. (1995). The fourth state of consciousness: The Thuriya
Avastha. Psychiatry and clinical neurosciences, 49(2), 107-110.
\item Ramey-Gassert, L. (1997). Learning science beyond the classroom. The
Elementary School Journal, 97(4), 433-450.
\item Reio Jr, T. G., Petrosko, J. M., Wiswell, A. K., \& Thongsukmag, J.
(2006). The measurement and conceptualization of curiosity. The Journal
of Genetic Psychology, 167(2), 117-135.
\item Reynolds, R. (1992). Super heroes: A modern mythology. Univ. Press
of Mississippi.
\item Riccati, J. (1724). Animadversiones in aequationes differentiales
secundi gradus. Actorum Eruditorum Supplementa, 8(1724), 66-73.
\item Robertson, E. M., Pascual-Leone, A., \& Press, D. Z. (2004). Awareness
modifies the skill-learning benefits of sleep. Current Biology, 14(3),
208-212.
\item Roy, R. K., Meszynska, A., Laure, C., Charles, L., Verchin, C., \&
Lutz, J. F. (2015). Design and synthesis of digitally encoded polymers
that can be decoded and erased. Nature communications, 6, 7237.
\item Rubinstein, M. E. (1973). A comparative statics analysis of risk premiums.
The Journal of Business, 46(4), 605-615.
\item Rubinstein, M. (1976). The valuation of uncertain income streams and
the pricing of options. The Bell Journal of Economics, 407-425.
\item Russell, S. J., \& Norvig, P. (1995). Artificial Intelligence: A Modern
Approach. Prentice-Hall, Englewood Cliffs, 25, 27.
\item Russell, S., Dewey, D., \& Tegmark, M. (2015). Research priorities
for robust and beneficial artificial intelligence. AI Magazine, 36(4),
105-114.
\item Samorini, G. (1992). The oldest representations of hallucinogenic
mushrooms in the world. Integration, 2, 3.
\item Sato, C., Takeuchi, S., \& Okude, N. (2011). Experience-based curiosity
model: Curiosity extracting model regarding individual experiences
of urban spaces. Design, User Experience, and Usability. Theory, Methods,
Tools and Practice, 635-644.
\item Saygin, A. P., Cicekli, I., \& Akman, V. (2000). Turing test: 50 years
later. Minds and machines, 10(4), 463-518.
\item Schmidhuber, J. (2015). Deep learning in neural networks: An overview.
Neural networks, 61, 85-117.
\item Schwartz, R. H., \& Smith, D. E. (1988). Hallucinogenic mushrooms.
Clinical pediatrics, 27(2), 70-73.
\item Scott, E. C. (2009). Evolution vs. creationism: An introduction (Vol.
62). Univ of California Press.
\item Searle, J. R. (1980). Minds, brains, and programs. Behavioral and
brain sciences, 3(3), 417-424.
\item Searle, J. R. (1982). The Chinese room revisited. Behavioral and brain
sciences, 5(2), 345-348.
\item Searle, J. R. (1990). Is the brain’s mind a computer program. Scientific
American, 262(1), 26-31.
\item Searle, J. (2001). Chinese Room Argument, The. Encyclopedia of cognitive
science.
\item Searle, J. R. (2004). Mind: a brief introduction. Oxford University
Press.
\item Shlens, J. (2014). A tutorial on principal component analysis. arXiv
preprint arXiv:1404.1100.
\item Simon, H. A. (1962). The Architecture of Complexity. Proceedings of
the American Philosophical Society, 106(6), 467-482.
\item Sipser, M. (2006). Introduction to the Theory of Computation (Vol.
2). Boston: Thomson Course Technology.
\item Smith, M. (1979). Behaviour of the Koala, Phascolarctos Cinereus Goldfuss,
in Captivity. 1. Non-Social Behaviour. Wildlife Research, 6(2), 117-129.
\item Snow, C. E., \& Hoefnagel-Höhle, M. (1978). The critical period for
language acquisition: Evidence from second language learning. Child
development, 1114-1128.
\item Soranzo, A., \& Epure, E. (2014). Very simply explicitly invertible
approximations of normal cumulative and normal quantile function.
Applied Mathematical Sciences, 8(87), 4323-4341.
\item Sorzano, C. O. S., Vargas, J., \& Montano, A. P. (2014). A survey
of dimensionality reduction techniques. arXiv preprint arXiv:1403.2877.
\item Stageberg, N. C. (1968). Structural Ambiguity for English Teachers.
In Selected Addresses Delivered at the Conference on English Education
(No. 6, pp. 29-34). National Council of Teachers of English.
\item Stein, C. M. (1973). Estimation of the mean of a multivariate normal
distribution. Proceedings of the Prague Symposium of Asymptotic Statistics.
\item Stein, C. M. (1981). Estimation of the mean of a multivariate normal
distribution. The annals of Statistics, 1135-1151.
\item Sternberg, R. J. (2018). Speculations on the role of Successful Intelligence
in solving contemporary world problems. Journal of Intelligence, 6(1),
4.
\item Storer, J. (1988). Data compression. Computer Science Press, Rockville,
Maryland.
\item Swanson, D. R. (1977). Information retrieval as a trial-and-error
process. The Library Quarterly, 47(2), 128-148.
\item Sutton, R. S., \& Barto, A. G. (1998). Reinforcement learning: An
introduction (Vol. 1, No. 1). Cambridge: MIT press.
\item Tahani, V. (1977). A conceptual framework for fuzzy query processing—a
step toward very intelligent database systems. Information Processing
\& Management, 13(5), 289-303.
\item Takeuchi, A., \& Amari, S. I. (1979). Formation of topographic maps
and columnar microstructures in nerve fields. Biological Cybernetics,
35(2), 63-72.
\item Taleb, N. N. (2007). The black swan: the impact of the highly improbable.
NY: Random House.
\item Teerapabolarn, K. (2013). Stein's identity for discrete distributions.
International Journal of Pure and Applied Mathematics, 83(4), 565.
\item Thompson, K. F., Gokler, C., Lloyd, S., \& Shor, P. W. (2016). Time
independent universal computing with spin chains: quantum plinko machine.
New Journal of Physics, 18(7), 073044.
\item Tou, F. N., Williams, M. D., Fikes, R., Henderson Jr, D. A., \& Malone,
T. W. (1982). RABBIT: An Intelligent Database Assistant. In AAAI (pp.
314-318).
\item Turing, A. M. (1950). Computing machinery and intelligence. Mind,
59(236), 433-460.
\item Vera, F. M., Manzaneque, J. M., Maldonado, E. F., Carranque, G. A.,
Rodriguez, F. M., Blanca, M. J., \& Morell, M. (2009). Subjective
sleep quality and hormonal modulation in long-term yoga practitioners.
Biological psychology, 81(3), 164-168.
\item Von Ahn, L., Blum, M., \& Langford, J. (2004). Telling humans and
computers apart automatically. Communications of the ACM, 47(2), 56-60.
\item Wagner, R. K., \& Sternberg, R. J. (1985). Practical intelligence
in real-world pursuits: The role of tacit knowledge. Journal of personality
and social psychology, 49(2), 436.
\item Waldfogel, S. (1948). The frequency and affective character of childhood
memories. Psychological Monographs: General and Applied, 62(4), i.
\item Weinberg, R. A. (1989). Intelligence and IQ: Landmark issues and great
debates. American Psychologist, 44(2), 98.
\item Williams, M. R. (1997). A history of computing technology. IEEE Computer
Society Press.
\item Wolfe, T. (2005). The right stuff. Random House.
\item Wonglimpiyarat, J. (2012). Technology strategies and standard competition—Comparative
innovation cases of Apple and Microsoft. The Journal of High Technology
Management Research, 23(2), 90-102.
\item Wooldridge, M., \& Jennings, N. R. (1995). Intelligent agents: Theory
and practice. The knowledge engineering review, 10(2), 115-152.
\item Yannakakis, G. N., \& Togelius, J. (2015). A panorama of artificial
and computational intelligence in games. IEEE Transactions on Computational
Intelligence and AI in Games, 7(4), 317-335.
\item Yang, M. (2008). Normal log-normal mixture, leptokurtosis and skewness.
Applied Economics Letters, 15(9), 737-742.
\item You, J. (2015). Beyond the turing test. Science, 347(6218), 116-116.
\item Young, H. P. (2009). Learning by trial and error. Games and economic
behavior, 65(2), 626-643.
\item Young, J. W. (1965). A Technique for Getting Ideas. CreateSpace.
\item Zhang, Q., Cheng, L., \& Boutaba, R. (2010). Cloud computing: state-of-the-art
and research challenges. Journal of internet services and applications,
1(1), 7-18.
\item Zogheib, B., \& Hlynka, M. (2009). Approximations of the Standard
Normal Distribution. University of Windsor, Department of Mathematics
and Statistics.
\end{doublespace}
\end{enumerate}
\begin{doublespace}

\section{\label{sec:From-Words-to-Symbols}Appendix: From Words to Symbols:
Mathematical Ingredients for a Curious and Confident Model of Intellect}
\end{doublespace}

\begin{doublespace}
\textit{The mathematical concepts discussed in this appendix are utilized
in Section (\ref{sec:A-Road-map-for}). Each sub-section below is
employed in different steps of the algorithm given in Section (\ref{sec:A-Road-map-for}).
Elaborate explanations, regarding how the below mathematical components
are necessary for the incubation of intelligence, are provided in
Section (\ref{sec:A-Road-map-for}) and the corresponding road-map
and also linked into the narrative throughout the article.}
\end{doublespace}
\begin{doublespace}

\subsection{Notation and Terminology for Key Results}
\end{doublespace}
\begin{itemize}
\begin{doublespace}
\item $D_{BC}\left(p_{i},p_{i}^{\prime}\right)$, the Bhattacharyya Distance
between two multinomial populations each consisting of $k$ categories
classes with associated probabilities $p_{1},p_{2},...,p_{k}$ and
$p_{1}^{\prime},p_{2}^{\prime},...,p_{k}^{\prime}$ respectively.
\item $\rho\left(p_{i},p_{i}^{\prime}\right)$, the Bhattacharyya Coefficient.
\item $D_{BC-N}(p,q)$ is the Bhattacharyya distance between $p$ and $q$
normal distributions or classes.
\item $D_{BC-MN}\left(p_{1},p_{2}\right)$ is the Bhattacharyya distance
between two multivariate normal distributions, $\boldsymbol{p_{1}},\boldsymbol{p_{2}}$
where $\boldsymbol{p_{i}}\sim\mathcal{N}(\boldsymbol{\mu}_{i},\,\boldsymbol{\Sigma}_{i})$.
\item $D_{BC-TN}(p,q)$ is the Bhattacharyya distance between $p$ and $q$
truncated normal distributions or classes.
\item $D_{BC-TMN}\left(p_{1},p_{2}\right)$ is the Bhattacharyya distance
between two truncated multivariate normal distributions, $\boldsymbol{p_{1}},\boldsymbol{p_{2}}$
where $\boldsymbol{p_{i}}\sim\mathcal{N}(\boldsymbol{\mu}_{i},\,\boldsymbol{\Sigma}_{i},\,\boldsymbol{a}_{i},\,\boldsymbol{b}_{i})$.
\item $F\left(t\right)$, is the installed base fraction with respect to
the adoption of a new product in a population.
\item $f\left(t\right)$, is the change of the installed base fraction or
the likelihood of purchase at time $t$ of a new product i.e. ${\displaystyle \ f(t)={\frac{d}{dt}}F(t)}$.
\item $p$, is the coefficient of innovation with respect to the adoption
of a new product in a population.
\item $q$, is the coefficient of imitation with respect to the adoption
of a new product in a population.
\item Sales (or new adopters) $S\left(t\right)$ at time $t$ is the rate
of change of installed base, that is, $f\left(t\right)$ multiplied
by the ultimate market potential $m$. 
\end{doublespace}
\end{itemize}
\begin{doublespace}

\subsection{\label{subsec:Bass-Model-of}Bass Model of Diffusion for Information
Accumulation}
\end{doublespace}

\begin{doublespace}
Collecting new pieces of information is the behavioral parallel we
draw to creating curiosity in our agents. One of the simplest forms
of the Bass model and also the original one from the pioneer (Bass
1969) can be written as,
\[
\frac{f\left(t\right)}{1-F\left(t\right)}=p+qF\left(t\right)
\]
\[
F\left(t\right)=\int_{0}^{t}f\left(u\right)du
\]
Here,

$f\left(t\right)$, is the change of the installed base fraction or
the likelihood of purchase at time $t$.

$F\left(t\right)$, is the installed base fraction.

$p$, is the coefficient of innovation.

$q$, is the coefficient of imitation.

Sales $S\left(t\right)$ at time $t$ is the rate of change of installed
base (i.e. adoption), that is, $f\left(t\right)$ multiplied by the
ultimate market potential $m$. This is given by,
\[
S\left(t\right)=mf\left(t\right)
\]
\[
S\left(t\right)=m\frac{\left(p+q\right)^{2}}{p}\frac{e^{-\left(p+q\right)t}}{\left(1+\frac{q}{p}e^{-\left(p+q\right)t}\right)^{2}}
\]

(Niu 2002) is a stochastic formulation of the Bass model of new product
diffusion. As alternatives, we could use models used in economics
for the spread of rumors and herd behavior. (Banerjee 1993) has a
discussion of information transmission processes, which for our purposes
are similar to information collection processes. (Banerjee 1992) a
sequential decision model in which each decision maker looks at the
decisions made by previous decision makers in taking her own decision.
This shows that the decision rules that are chosen by optimizing individuals
will be characterized by herd behavior, i.e., people will be doing
what others are doing rather than using their information.
\end{doublespace}
\begin{doublespace}

\subsection{\label{subsec:Knowledge-Store}Knowledge Store}
\end{doublespace}

\begin{doublespace}
We could use the developments in the field of text parsing and storing
(Piskorski \& Neumann 2000; Jacobs 2014), to create keyword based
database(s) to hold bits of learning that the agent has gathered.
The knowledge store has to be processed periodically to establish
and reestablish the connections between the different stored elements.
This feature would be of assistance in being able to recollect what
has been learnt. The connections are established based on the Bhattacharyya
distance, discussed next, and when appropriate we use dimension reduction
techniques so that this distance measure could be applied. For intelligent
database systems and related query developments, see: (Tahani 1977;
Tou, et al., 1982; Parsaye \& Chignell 1993; Catania \& Zarri 2000).
\end{doublespace}
\begin{doublespace}

\subsection{\label{subsec:Bhattacharyya-Distance}Bhattacharyya Distance for
Information Comparison}
\end{doublespace}

\begin{doublespace}
We use the Bhattacharyya distance (Bhattacharyya 1943, 1946) as a
measure of similarity or dissimilarity between the probability distributions
of the two entities we are looking to compare. These entities could
be two information sources, two securities, groups of securities,
markets or any statistical populations that we are interested in studying
(Kashyap 2019). The Bhattacharyya distance is defined as the negative
logarithm of the Bhattacharyya coefficient. 
\[
D_{BC}\left(p_{i},p_{i}^{\prime}\right)=-\ln\left[\rho\left(p_{i},p_{i}^{\prime}\right)\right]
\]
The Bhattacharyya coefficient is calculated as shown below for discrete
and continuous probability distributions. 
\[
\rho\left(p_{i},p_{i}^{\prime}\right)=\sum_{i}^{k}\sqrt{p_{i}p_{i}^{\prime}}
\]
\[
\rho\left(p_{i},p_{i}^{\prime}\right)=\int\sqrt{p_{i}\left(x\right)p_{i}^{\prime}\left(x\right)}dx
\]

Bhattacharyya’s original interpretation of the measure was geometric
(Derpanis 2008). He considered two multinomial populations each consisting
of $k$ category classes with associated probabilities $p_{1},p_{2},...,p_{k}$
and $p_{1}^{\prime},p_{2}^{\prime},...,p_{k}^{\prime}$ respectively.
Then, as $\sum_{i}^{k}p_{i}=1$ and $\sum_{i}^{k}p_{i}^{\prime}=1$,
he noted that $(\sqrt{p_{1}},...,\sqrt{p_{k}})$ and $(\sqrt{p_{1}^{\prime}},...,\sqrt{p_{k}^{\prime}})$
could be considered as the direction cosines of two vectors in $k-$dimensional
space referred to a system of orthogonal co-ordinate axes. As a measure
of divergence between the two populations Bhattacharyya used the square
of the angle between the two position vectors. If $\theta$ is the
angle between the vectors then,
\[
\rho\left(p_{i},p_{i}^{\prime}\right)=cos\theta=\sum_{i}^{k}\sqrt{p_{i}p_{i}^{\prime}}
\]
Thus if the two populations are identical, $cos\theta=1$ corresponding
to $\theta=0$. Hence we see the intuitive motivation behind the definition
as the vectors are co-linear. Bhattacharyya further showed that, by
passing to the limiting case, a measure of divergence could be obtained
between two populations defined in any way given that the two populations
have the same number of variates. The value of coefficient then lies
between $0$ and $1$. 
\[
0\leq\rho\left(p_{i},p_{i}^{\prime}\right)\leq=1
\]
\[
0\leq D_{BC}\left(p_{i},p_{i}^{\prime}\right)\leq\infty
\]
We get the following formulae (Lee \& Bretschneider 2012) for the
Bhattacharyya distance when applied to the case of two uni-variate
normal distributions. 
\[
D_{BC-N}(p,q)=\frac{1}{4}\ln\left(\frac{1}{4}\left(\frac{\sigma_{p}^{2}}{\sigma_{q}^{2}}+\frac{\sigma_{q}^{2}}{\sigma_{p}^{2}}+2\right)\right)+\frac{1}{4}\left(\frac{(\mu_{p}-\mu_{q})^{2}}{\sigma_{p}^{2}+\sigma_{q}^{2}}\right)
\]

$\sigma_{p}$ is the variance of the $p-$th distribution, 

$\mu_{p}$ is the mean of the $p-$th distribution, and 

$p,q$ are two different distributions.

The original paper on the Bhattacharyya distance (Bhattacharyya 1943)
mentions a natural extension to the case of more than two populations.
For an $M$ population system, each with $k$ random variates, the
definition of the coefficient becomes, 
\[
\rho\left(p_{1},p_{2},...,p_{M}\right)=\int\cdots\int\left[p_{1}\left(x\right)p_{2}\left(x\right)...p_{M}\left(x\right)\right]^{\frac{1}{M}}dx_{1}\cdots dx_{k}
\]

For two multivariate normal distributions, $\boldsymbol{p_{1}},\boldsymbol{p_{2}}$
where $\boldsymbol{p_{i}}\sim\mathcal{N}(\boldsymbol{\mu}_{i},\,\boldsymbol{\Sigma}_{i}),$
\[
D_{BC-MN}\left(p_{1},p_{2}\right)=\frac{1}{8}(\boldsymbol{\mu}_{1}-\boldsymbol{\mu}_{2})^{T}\boldsymbol{\Sigma}^{-1}(\boldsymbol{\mu}_{1}-\boldsymbol{\mu}_{2})+\frac{1}{2}\ln\,\left(\frac{\det\boldsymbol{\Sigma}}{\sqrt{\det\boldsymbol{\Sigma}_{1}\,\det\boldsymbol{\Sigma}_{2}}}\right),
\]

$\boldsymbol{\mu}_{i}$ and $\boldsymbol{\Sigma}_{i}$ are the means
and covariances of the distributions, and $\boldsymbol{\Sigma}=\frac{\boldsymbol{\Sigma}_{1}+\boldsymbol{\Sigma}_{2}}{2}$.
We need to keep in mind that a discrete sample could be stored in
matrices of the form $A$ and $B$, where, $n$ is the number of observations
and $m$ denotes the number of variables captured by the two matrices.
\[
\boldsymbol{A_{m\times n}}\sim\mathcal{N}\left(\boldsymbol{\mu_{1}},\boldsymbol{\varSigma_{1}}\right)
\]
\[
\boldsymbol{B_{m\times n}}\sim\mathcal{N}\left(\boldsymbol{\mu_{2}},\boldsymbol{\varSigma_{2}}\right)
\]

\end{doublespace}
\begin{doublespace}

\subsection{\label{subsec:Dimension-Reduction}Dimension Reduction before Information
Comparison}
\end{doublespace}

\begin{doublespace}
A key requirement to apply the Bhattacharyya distance in practice
is to have data-sets with the same number of dimensions. (Fodor 2002;
Burges 2009; Sorzano, Vargas \& Montano 2014) are comprehensive collections
of methodologies aimed at reducing the dimensions of a data-set using
Principal Component Analysis or Singular Value Decomposition and related
techniques. (Johnson \& Lindenstrauss 1984) proved a fundamental result
(JL Lemma) that says that any $n$ point subset of Euclidean space
can be embedded in $k=O(log\frac{n}{\epsilon^{2}})$ dimensions without
distorting the distances between any pair of points by more than a
factor of $\left(1\pm\epsilon\right)$, for any $0<\epsilon<1$. Whereas
principal component analysis is only useful when the original data
points are inherently low dimensional, the JL Lemma requires absolutely
no assumption on the original data. Also, note that the final data
points have no dependence on $d$, the dimensions of the original
data which could live in an arbitrarily high dimension. We use the
version of the bounds for the dimensions of the transformed subspace
given in (Frankl \& Maehara 1988; 1990; Dasgupta \& Gupta 1999).
\end{doublespace}
\begin{lem}
\begin{doublespace}
\label{Prop:Johnson and Lindenstrauss --- Dasgupta and Gupta}For
any $0<\epsilon<1$ and any integer $n$, let $k$ be a positive integer
such that 
\[
k\geq4\left(\frac{\epsilon^{2}}{2}-\frac{\epsilon^{3}}{3}\right)^{-1}\ln n
\]
Then for any set $V$ of $n$ points in $\boldsymbol{R}^{d}$, there
is a map $f:\boldsymbol{R}^{d}\rightarrow\boldsymbol{R}^{k}$ such
that for all $u,v\in V$, 
\[
\left(1-\epsilon\right)\Vert u-v\Vert^{2}\leq\Vert f\left(u\right)-f\left(v\right)\Vert^{2}\leq\left(1+\epsilon\right)\Vert u-v\Vert^{2}
\]
Furthermore, this map can be found in randomized polynomial time and
one such map is $f=\frac{1}{\sqrt{k}}Ax$ where, $x\in\boldsymbol{R}^{d}$
and $A$ is a $k\times d$ matrix in which each entry is sampled i.i.d
from a Gaussian $N\left(0,1\right)$ distribution.
\end{doublespace}
\end{lem}
\begin{doublespace}
(Kashyap 2019) provides expressions for the density functions after
dimension transformation when considering log normal distributions,
truncated normal and truncated multivariate normal distributions (Norstad
1999; Horrace 2005; Kiani, etal 2008; Yang 2008; Burkardt 2014). These
results are applicable in the context of many variables observed in
real life such as stock prices, heart rates, inventory levels, and
volatilities, which do not take on negative values. We also require
the expression for the dimension transformed normal distribution.
Techniques for numerical approximations are useful since the normal
cumulative distribution (Zogheib \& Hlynka 2009; Soranzo \& Epure
2014) is a better candidate to model a reward variable, which could
take on negative values. Error function approximations are also helpful
choices (Cody 1969; Chiani, Dardari \& Simon 2003). A relationship
between co-variance (Stein 1973; 1981; Kimeldorf \& Sampson 1973;
Rubinstein 1973; 1976; Kattumannil 2009; Teerapabolarn 2013) and distance
measures is also derived. We point out that these mathematical concepts
have many uses outside the domain of artificial intelligence.
\end{doublespace}

\end{document}